\RequirePackage{fix-cm}
\documentclass[smallextended]{svjour3}       
\smartqed  
\usepackage{amsmath,amssymb,amsfonts}%
\usepackage{mathrsfs}%
\usepackage[title]{appendix}%
\usepackage{textcomp}%
\usepackage{manyfoot}%
\usepackage{booktabs}%
\usepackage{algorithm}%
\usepackage{algorithmicx}%
\usepackage{algpseudocode}%
\usepackage{listings}%
\usepackage{listings}
\usepackage{pifont}
\usepackage[dvipsnames]{xcolor}
\usepackage{graphicx}%
\usepackage{url}
\usepackage{multicol}

\makeatletter

\newcommand\PrologPredicateStyle{}
\newcommand\PrologVarStyle{}
\newcommand\PrologAnonymVarStyle{}
\newcommand\PrologAtomStyle{}
\newcommand\PrologOtherStyle{}
\newcommand\PrologCommentStyle{}

\newif\ifpredicate@prolog@
\newif\ifwithinparens@prolog@

\lst@SaveOutputDef{`_}\underscore@prolog

\newcount\currentchar@prolog

\newcommand\@testChar@prolog%
{%
  \ifnum\lst@mode=\lst@Pmode%
    \detectTypeAndHighlight@prolog%
  \else
    \ifwithinparens@prolog@%
      \detectTypeAndHighlight@prolog%
    \fi
  \fi
  \global\predicate@prolog@false%
}

\newcommand\detectTypeAndHighlight@prolog
{%
  \def\lst@thestyle{\PrologAtomStyle}%
  \ifpredicate@prolog@%
    \def\lst@thestyle{\PrologPredicateStyle}%
  \else
    \expandafter\splitfirstchar@prolog\expandafter{\the\lst@token}%
    \expandafter\ifx\@testChar@prolog\underscore@prolog%
      \ifnum\lst@length=1%
        \let\lst@thestyle\PrologAnonymVarStyle%
      \else
        \let\lst@thestyle\PrologVarStyle%
      \fi
    \else
      \currentchar@prolog=65
      \loop
        \expandafter\ifnum\expandafter`\@testChar@prolog=\currentchar@prolog%
          \let\lst@thestyle\PrologVarStyle%
          \let\iterate\relax
        \fi
        \advance \currentchar@prolog by 1
        \unless\ifnum\currentchar@prolog>90
      \repeat
    \fi
  \fi
}
\newcommand\splitfirstchar@prolog{}
\def\splitfirstchar@prolog#1{\@splitfirstchar@prolog#1\relax}
\newcommand\@splitfirstchar@prolog{}
\def\@splitfirstchar@prolog#1#2\relax{\def\@testChar@prolog{#1}}

\def\beginlstdelim#1#2%
{%
  \def\endlstdelim{\PrologOtherStyle #2\egroup}%
  {\PrologOtherStyle #1}%
  \global\predicate@prolog@false%
  \withinparens@prolog@true%
  \bgroup\aftergroup\endlstdelim%
}

\newcommand\lang@prolog{Prolog-pretty}
\expandafter\lst@NormedDef\expandafter\normlang@prolog%
  \expandafter{\lang@prolog}

\expandafter\expandafter\expandafter\lstdefinelanguage\expandafter%
{\lang@prolog}
{%
  language            = Prolog,
  keywords            = {},      %
  showstringspaces    = false,
  alsoletter          = (,
  moredelim           = **[is][\beginlstdelim{(}{)}]{(}{)},
  MoreSelectCharTable =
    \lst@DefSaveDef{`(}\opparen@prolog{\global\predicate@prolog@true\opparen@prolog},
}

\newcommand\@ddedToOutput@prolog\relax
\lst@AddToHook{Output}{\@ddedToOutput@prolog}

\lst@AddToHook{PreInit}
{%
  \ifx\lst@language\normlang@prolog%
    \let\@ddedToOutput@prolog\@testChar@prolog%
  \fi
}

\lst@AddToHook{DeInit}{\renewcommand\@ddedToOutput@prolog{}}

\makeatother

\definecolor{PrologVar}{RGB}{65,105,225}%
\definecolor{PrologPredicate} {RGB}{205,92,92}%
\definecolor{PrologAnonymVar}{RGB}{000,127,000}
\definecolor{PrologAtom}     {RGB}{95,95,95}
\definecolor{PrologComment}  {RGB}{063,128,127}
\definecolor{PrologOther}    {RGB}{000,000,000}
\definecolor{backcolour}{rgb}{0.96,0.96,0.94}

\renewcommand\PrologPredicateStyle{\color{PrologPredicate}}
\renewcommand\PrologVarStyle{\color{PrologVar}}
\renewcommand\PrologAnonymVarStyle{\color{PrologAnonymVar}}
\renewcommand\PrologAtomStyle{\color{PrologAtom}}
\renewcommand\PrologCommentStyle{\itshape\color{PrologComment}}
\renewcommand\PrologOtherStyle{\color{PrologOther}}

\lstdefinestyle{Prolog-pygsty}
{
  language     = Prolog-pretty,
  upquote      = true,
backgroundcolor=\color{backcolour},   
  stringstyle  = \PrologAtomStyle,
  commentstyle = \PrologCommentStyle,
  literate     =
    {:-}{{\PrologOtherStyle :-}}2
    {,}{{\PrologOtherStyle ,}}1
    {.}{{\PrologOtherStyle .}}1
}

\newcommand{\cmark}{\ding{51}}%
\newcommand{\xmark}{\ding{55}}%
\newcommand{\CMARK}{\textcolor{ForestGreen}{\cmark}}
\newcommand{\XMARK}{\textcolor{BrickRed}{\xmark}}
\newcommand{\eg}{\emph{e.g.}~} 
\newcommand{\ie}{\emph{i.e.}~} {}
\newcommand{\cf}{\emph{cf.}~} 

\newcommand{\clevrobj}[1]{\vcenter{\hbox{\includegraphics[height=10pt]{#1}}}}

\begin{document}

\title{Learning Differentiable Logic Programs for Abstract Visual Reasoning
}


\author{Hikaru Shindo         \and
         Viktor Pfanschilling \and
         Devendra Singh Dhami \and
         Kristian Kersting
}


\institute{Hikaru Shindo \at
              \email{hikaru.shindo@tu-darmstadt.de}           
}

\date{Accepted: Aug 4, 2024}

\maketitle

\abstract{Visual reasoning is essential for building intelligent agents that understand the world and perform problem-solving beyond perception. Differentiable forward reasoning has been developed to integrate reasoning with gradient-based machine learning paradigms. 
However, due to the memory intensity, most existing approaches do not bring the best of the expressivity of first-order logic, excluding a crucial ability to solve \emph{abstract visual reasoning}, where agents need to perform reasoning by using analogies on abstract concepts in different scenarios. 
To overcome this problem, we propose \emph{NEUro-symbolic Message-pAssiNg reasoNer (NEUMANN)}, which is a graph-based differentiable forward reasoner, passing messages in a memory-efficient manner and handling structured programs with functors.
Moreover, we propose a computationally-efficient structure learning algorithm to perform explanatory program induction on complex visual scenes.
To evaluate, in addition to conventional visual reasoning tasks, we propose a new task, \emph{visual reasoning behind-the-scenes}, where agents need to learn abstract programs and then answer queries by imagining scenes that are not observed. 
We empirically demonstrate that NEUMANN solves visual reasoning tasks efficiently, outperforming neural, symbolic, and neuro-symbolic baselines.}

\section{Introduction}
Deep Neural Networks (DNNs) are attracting considerable interest due to significant performances in crucial tasks in Artificial Intelligence~\cite{lecun2015deep} such as image recognition~\cite{Krizhevsky12imagenet}, game playing~\cite{Silver16alphago}, protein-structure prediction~\cite{AlphaFold2021}, and language modeling~\cite{brown2020language} to name a few. DNNs are essentially data-driven, \ie they perform pattern recognition statistically given data and perform prediction on new examples.
However, a critical gap exists between human intelligence and the current data-driven machine-learning paradigm. Humans can explain and understand what they see, imagine things they could see but have not yet, and perform planning to solve problems~\cite{Lake17buildingmachines}. Moreover, humans can learn from a small number of experiences~\cite{Tenenbaum11ScienceGrowMind,Silver2020fewexamples}, but DNNs such as transformers~\cite{Vaswani17attention,dwivedi2021transformergeneralization,zhao2020transformer-xh,devlin-etal-2019-bert,yun19graphtransformer} require a large dataset to achieve good performance on a specific task~\cite{Geman1992NeuralNetworksBiasVariance}. These essential intelligent aspects of humans, called \emph{model building}~\cite{Lake17buildingmachines}, are vital for human-level intelligence.

Logic has been a fundamental element of AI for providing knowledge representations and reasoning capabilities~\cite{Davis93WhatisKR,Russel09}. 
Inductive Logic Programming (ILP)~\cite{Muggleton91,Nienhuys97,Cropper20} is a framework to learn logic programs 
 given examples.
In stark contrast to DNNs, ILP gains some crucial advantages, \eg it can learn from small data, and it can learn explicit programs, which are interpretable by humans.
Recently, Differentiable ILP ($\partial$ILP) has been proposed~\cite{Evans18}, where they perform gradient-based learning of logic programs. In $\partial$ILP, \emph{forward reasoning}, which derives all possible consequences given logic programs, is implemented using only differentiable operations by encoding logic programs into \emph{tensors}. Thus it can be easily combined with DNNs for the perception and perform ILP on visual inputs.
However, tensor-based differentiable forward reasoning is memory-intensive. Thus it assumes that logic programs to be handled are simple, \eg each predicate takes at most two arguments, each clause has at most two body atoms, and no functors are allowed. 
$\partial$ILP-ST~\cite{Shindo21} has been developed to deal with structured logic programs with functors in $\partial$ILP, leading to $\alpha$ILP~\cite{shindo23alphailp}, which can learn classification rules on complex visual scenes. 
They address the memory-consumption problem by performing a beam-search over clauses instead of generating all possible clauses by templates.
However, performing a beam search is computationally expensive because every candidate of clauses needs to be evaluated in each step. 
Thus it takes longer to complete when handling complex programs and does not scale for more challenging tasks where agents play multiple roles, \eg understanding visual scenes, learning abstract operations and solving queries by abstract reasoning.

To mitigate this issue, we develop a memory-efficient differentiable forward reasoner and a computationally efficient learning strategy. We propose \emph{NEUro-symbolic Message-pAssiNg reasoNer (NEUMANN)}, a graph-based approach for differentiable forward reasoning, sending messages in a memory-efficient manner. 
We first introduce a new graph-based representation of logic programs in first-order logic and then perform differentiable reasoning via message passing.
The graph structure efficiently encodes the reasoning process by connecting logical atoms.
Then, we propose a computationally-efficient learning algorithm for NEUMANN by combining gradient-based scoring and differentiable sampling.
Instead of scoring each clause exactly to perform a beam search, NEUMANN computes gradients over candidate clauses for a classification loss and uses them as approximated scores to generate new clauses. 
By doing so, NEUMANN avoids nested scoring loops over clauses, which has been a computational bottleneck of the beam-search approach.

\begin{figure}[t]
\centering
\includegraphics[width=\linewidth]{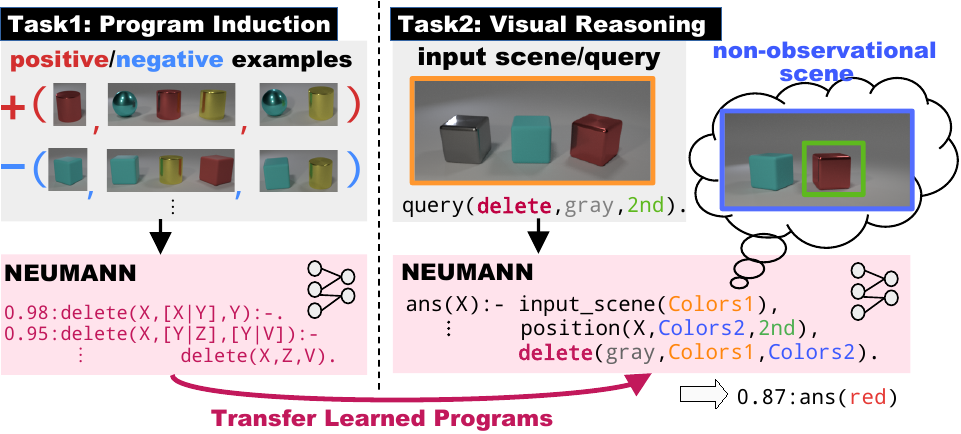}
\caption{\textbf{Reasoning behind the scenes}:  The goal of this task is to compute the answer to a query, \eg  \emph{``What is the color of the second left-most object after deleting a gray object?"} given a visual scene. To answer this query, the agent needs to reason behind the scenes and understand abstract operations on objects. \textbf{(Task 1, left)} In the first task, the agent needs to induce an explicit program given visual examples, where each example consists of several visual scenes that describe the input and the output of the operation to be learned. The abstract operations can be described by first-order logic with functors. 
\textbf{(Task 2, right)} In the second task, the agent needs to apply the learned programs to new situations to solve queries about non-observational scenes. (Best viewed in color)
}
\label{fig:behind-the-scene}
\end{figure}

The memory-efficient reasoning and computationally-efficient learning enable NEUMANN to solve \emph{abstract visual reasoning}, where the agent needs to perform reasoning by using analogies on abstract concepts in different scenarios. 
To evaluate this, we propose a new task, \emph{Visual Reasoning Behind the Scenes}, where the agent needs to perform complex visual reasoning imagining scenes that are not observed.
Fig.~\ref{fig:behind-the-scene} illustrates a Behind-the-Scenes task whose goal is to compute the answer of a query, \eg  \emph{``What is the color of the second left-most object after deleting a gray object?"} given a visual scene. 
In turn, it consists of two sub-tasks. 
The first is to induce abstract programs from visual scenes, \eg \emph{deletion} of objects, as shown in the left of Fig.~\ref{fig:behind-the-scene}. 
The second is to solve the queries where the answers are derived by reasoning about non-observational scenes.
To solve, the agent needs to learn abstract operations from visual input and perform efficient reasoning. 
The task assesses the following four essential model-building capacities: (1) learning from a small number of examples, (2) understanding complex visual scenes deeply, (3) learning explanatory programs to transfer to new tasks, and (4) imagining situations that have not been observed directly. Behind-the-Scenes is the first benchmark to cover all of these four aspects. We highlight on Tab.~\ref{tab:compare_tasks} the difference from previous visual reasoning tasks in these aspects. 
Behind-the-Scenes serves as a legitimate task and dataset for the model-building abilities, which is beneficial to foster the machine-learning paradigm to perform problem-solving beyond pattern recognition.

To summarize, we make the following important contributions:
\begin{enumerate}
    \item We propose NEUMANN\footnote{Code is available: \url{https://github.com/ml-research/neumann}}, a memory-efficient differentiable forward reasoner using message-passing. We theoretically and empirically show that NEUMANN requires less memory than conventional tensor-based differentiable forward reasoners~\cite{Evans18,Shindo21,shindo23alphailp}. Given $G$ ground atoms and $C^*$ ground clauses, conventional differentiable forward reasoners consume memory quadratically $\mathcal{O}(G \times C^*)$, but NEUMANN consumes linearly $\mathcal{O}(G + C^*)$. 
    \item We propose a computationally-efficient learning algorithm for NEUMANN to learn complex programs from visual scenes. NEUMANN performs gradient-based scoring and differentiable sampling, avoiding nested loops for scoring candidate clauses.
    \item We propose a new challenging task and a dataset, \emph{Visual Reasoning Behind the Scenes}, where the agents need to perform abstract visual learning and reasoning on complex visual scenes. The task requires the agents to learn abstract operations from small data on visual scenes and reason about non-observational scenes to answer queries. The task evaluates machine-learning models on the different essential model-building properties of intelligence beyond perception, which are not covered by the previously addressed visual reasoning benchmarks.
    \item  We empirically show that NEUMANN solves visual reasoning tasks such as Kandinsky patterns~\cite{Muller21kandinsky} and CLEVR-Hans~\cite{Stammer21} using less memory than conventional differentiable forward reasoners, outperforming neural baselines. More importantly, we show that NEUMANN efficiently solves the proposed Behind-the-Scenes task, outperforming conventional differentiable forward reasoners. To this end, we show that NEUMANN gains the advantages of scalable and explainable visual reasoning and learning against symbolic and neuro-symbolic baselines.
\end{enumerate}

\begin{table}[t]
\centering

\caption{\textbf{Comparison between Behind-the-Scenes and other visual reasoning benchmarks.} The Behind-the-scenes task assesses the four essential \emph{model-building} features: \textbf{(small data)} the task requires the model to learn from a small number of training data, \textbf{(visual scenes)} the task requires to handle complex visual scenes where several objects appear, \textbf{(explanatory)} the task requires to learn explanatory programs, and \textbf{(imagination)} the task requires answers obtained by reasoning about non-observational scenes.}
\begin{tabular}{lcccc}
                  & small data & visual scenes & explanatory & ``imagination" \\ \hline
VQA~\cite{Antol15vqa}               & \XMARK           & \CMARK             & \XMARK                 & \XMARK           \\
VQAR~\cite{Huang21scallop}             & \XMARK           & \CMARK             & \CMARK                 & \XMARK           \\
CLEVR~\cite{Johnson17}           & \XMARK           & \CMARK             & \XMARK                 & \XMARK           \\
CLEVRER~\cite{Yi2020CLEVRER}           & \XMARK           & \CMARK             & \XMARK                 & \CMARK           \\
CLEVR-Hans~\cite{Stammer21}      & \XMARK           & \CMARK             & \CMARK                 &  \XMARK          \\
MNIST-Addition~\cite{Manhaeve19}    & \XMARK          &  \XMARK             & \CMARK                 & \XMARK            \\
RAVEN~\cite{raven1998raven} & \CMARK & \CMARK & \XMARK & \XMARK \\
KandinskyPattern~\cite{Muller21kandinsky} & \CMARK           & \CMARK             & \CMARK                 & \XMARK           \\
Behind-the-Scenes & \CMARK          & \CMARK             & \CMARK                 & \CMARK           \\ \hline
\end{tabular}
\label{tab:compare_tasks}
\end{table}

\section{First-Order Logic, Differentiable Reasoning, and Graph Neural Networks}
Before introducing NEUMANN, we revisit the basic concepts of first-order logic and graph neural networks.

{\bf First-Order Logic (FOL).}
A {\it Language} $\mathcal{L}$ is a tuple $(\mathcal{P}, \mathcal{A}, \mathcal{F}, \mathcal{V})$,
where $\mathcal{P}$ is a set of predicates, $\mathcal{A}$ is a set of constants, $\mathcal{F}$ is a set of function symbols (functors), and $\mathcal{V}$ is a set of variables.
A {\it term} is a constant, a variable, or a term that consists of a functor.
A {\it ground term} is a term with no variables.
We denote $n$-ary predicate ${\tt p}$ by ${\tt p}/n$.
An {\it atom} is a formula ${\tt p(t_1, \ldots, t_n) }$, where ${\tt p}$ is an $n$-ary predicate symbol and ${\tt t_1, \ldots, t_n}$ are terms.
A {\it ground atom} or simply a {\it fact} is an atom with no variables.
A {\it literal} is an atom or its negation.
A {\it positive literal} is just an atom. 
A {\it negative literal} is the negation of an atom.
A {\it clause} is a finite disjunction ($\lor$) of literals. 
A {\it ground clause} is a clause with no variables.
A {\it definite clause} is a clause with exactly one positive literal.
If  $A, B_1, \ldots, B_n$ are atoms, then $ A \lor \lnot B_1 \lor \ldots \lor \lnot B_n$ is a definite clause.
We write definite clauses in the form of $A~\mbox{:-}~B_1,\ldots,B_n$.
Atom $A$ is called the {\it head}, and set of negative atoms $\{B_1, \ldots, B_n\}$ is called the {\it body}.
We call definite clauses by clauses for simplicity in this paper.
We denote $\mathit{true}$ as $\top$ and $\mathit{false}$ as $\bot$.
Substitution $\theta = \{\tt X_1 = t_1, ..., X_n = t_n\}$ is an assignment of term ${\tt t_i}$ to variable ${\tt X_i}$. An application of substitution $\theta$ to atom $A$ is written as $A \theta$. An atom is an atomic \emph{formula}. For formula $F$ and $G$, $\lnot F$, $F \land G$, and $F \lor G$ are also formulas.
\emph{Interpretation} of language $\mathcal{L}$ is a tuple $(\mathcal{D}, \mathcal{I}_\mathcal{A}, \mathcal{I}_\mathcal{F}, \mathcal{I}_\mathcal{P})$, 
where $\mathcal{D}$ is the  domain, $\mathcal{I}_\mathcal{A}$ is the assignments of an element in $\mathcal{D}$ for each constant ${\tt a} \in \mathcal{A}$,
$\mathcal{I}_\mathcal{F}$ is the assignments of a function from $\mathcal{D}^n$ to $\mathcal{D}$ for each $n$-ary function symbol ${\tt f} \in \mathcal{F}$, 
and $\mathcal{I}_\mathcal{P}$ is the assignments of a function from $\mathcal{D}^n$ to $\{ \top, \bot \}$ for each $n$-ary predicate ${\tt p} \in \mathcal{P}$.
For language $\mathcal{L}$ and formala $X$, an interpretation $\mathcal{I}$ is a \emph{model} if the truth value of $X$ w.r.t $\mathcal{I}$ is true.
Formula $X$ is a \emph{logical consequence} or \emph{logical entailment} of a set of formulas $\mathcal{H}$, denoted $\mathcal{H} \models X$, if, $\mathcal{I}$ is a model for $\mathcal{H}$ implies that $\mathcal{I}$ is a model for $X$ for every interpretation $\mathcal{I}$ of $\mathcal{L}$.

\textbf{(Differentiable) Forward Reasoning} is a data-driven approach of reasoning in FOL~\cite{Russel09}. 
Forward reasoning is performed by applying a function called the \emph{$T_\mathcal{C}$ operator}, deducing new ground atoms using given clauses and ground atoms.
For a set of clauses $\mathcal{C}$, $T_\mathcal{C}$ operator~\cite{Lloyd84} is a function that applies clauses in  $\mathcal{C}$ using given ground atoms $\mathcal{G}$, \ie  
\begin{equation}
  T_\mathcal{C}(\mathcal{G})=\mathcal{G} \cup \left\{ A ~\middle|~
  \begin{aligned} 
    &A ~\mbox{:-}~ B_1, \ldots, B_n \in \mathcal{C}^*  \\
     &(\{ B_1 \ldots, B_n \} \subseteq \mathcal{G})
  \end{aligned}
  \right\},
  \label{eq:T_c}
\end{equation}
where $\mathcal{C}^*$ is a set of all ground clauses that can be produced from $\mathcal{C}$.
\noindent
Note that the union with $\mathcal{G}$ is computed to hold the ground atoms in the previous steps.
The forward reasoning function can then be defined as a function that repeatedly applies the $T_\mathcal{C}$ operator to given ground atoms.

Differentiable forward reasoning~\cite{Evans18,Shindo21,shindo23alphailp} uses only simple tensor operations to compute forward reasoning. 
Given $G$ ground atoms and $C$ clauses, the reasoner computes the grounding of clauses, \ie removing variables, producing $C^*$ ground clauses, then it builds \emph{index tensor} $\mathbf{I} \in \mathbb{N}^{G \times C^*}$, which holds the indices of ground atoms for each ground clause. The differentiable forward reasoner computes logical entailment referring to the index tensor repeatedly.

\textbf{Graph Neural Networks.}
Graph Neural Network (GNN)~\cite{Scarselli09,Li16,kipf2017semi,Hamilton17,Schlichtkrull18gnn}  is a type of neural network that processes graphs as inputs.
An input data is represented as $(\mathsf{G}, \mathbf{x}_\mathit{node}, \mathbf{x}_\mathit{edge})$, where $\mathsf{G}$ is a directed or undirected graph, $\mathbf{x}_\mathit{node}$ represents node features, and $\mathbf{x}_\mathit{edge}$ represents edge features.
Given an input, GNN computes the node representations by performing message-passing:
\begin{align}
    x_i^{(t+1)} = f_\mathit{update}\left(x_i^{(t)}, \bigoplus\nolimits_{j \in \mathcal{N}(i)} c_{ji} \cdot x_j^{(t)} \right),
\end{align}
where $t \in \mathbb{N}$ is a time step, $x_i^{(t)}$ is a node feature of node $\mathsf{x_i}$ at time step $t$, $\mathcal{N}(i)$ is a set of indices of neighbors of node $\mathsf{x_i}$, $c_{ji}$ is an edge feature, and  $\bigoplus$ is an aggregation function to aggregate messages from neighbors.

\section{NEUMANN}
\begin{figure}
    \centering
    \includegraphics[width=.9\linewidth]{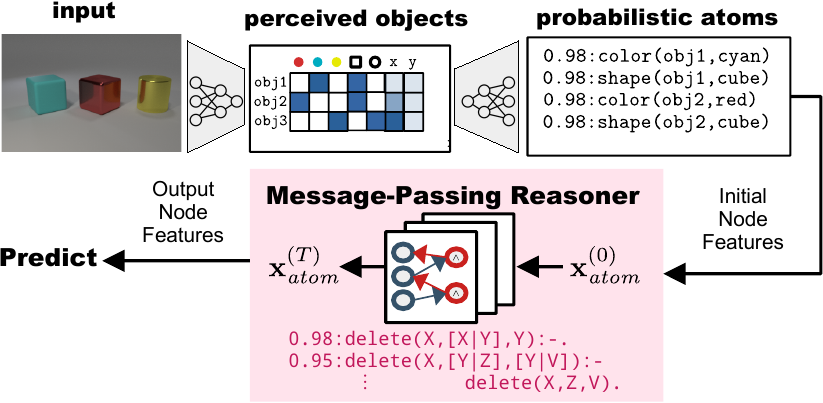}
    \caption{\textbf{The reasoning architecture in NEUMANN.}  Raw input images are fed into the visual-perception model. The output is converted into a set of probabilistic ground atoms.
    Differentiable forward reasoning is performed by a bi-directional message-passing algorithm. Logical entailment is computed softly using weighted clauses and probabilistic ground atoms. (Best viewed in color)}
    \label{fig:predict}
\end{figure}

NEUMANN computes logical entailment in a differentiable manner given visual input and weighted clauses.
Fig.~\ref{fig:predict} illustrates the overview of NEUMANN's reasoning pipeline. In contrast to conventional differentiable forward reasoners~\cite{Evans18,Shindo21,shindo23alphailp}, NEUMANN performs message-passing on graphs in the following steps:
\textbf{(Step 1)} A visual input is fed into a neural network to perceive objects in the scene.
The output of the neural network is encoded into a set of probabilistic atoms $\mathbf{x}_\mathit{atom}^{(0)}$.
\textbf{(Step 2)} 
Given input probabilistic atoms $\mathbf{x}_\mathit{atom}^{(0)}$, NEUMANN performs $T$ bi-directional message-passing steps.
The graph represents a set of weighted clauses, and 
the output node features $\mathbf{x}_\mathit{atom}^{(T)}$ represent probabilistic values of logical entailment given $\mathbf{x}_\mathit{atom}^{(0)}$ and weighted clauses. We describe each step in detail.

\subsection{Forward Reasoning Graph}
\begin{figure}[t]
    \centering
    \includegraphics[width=.75\linewidth]{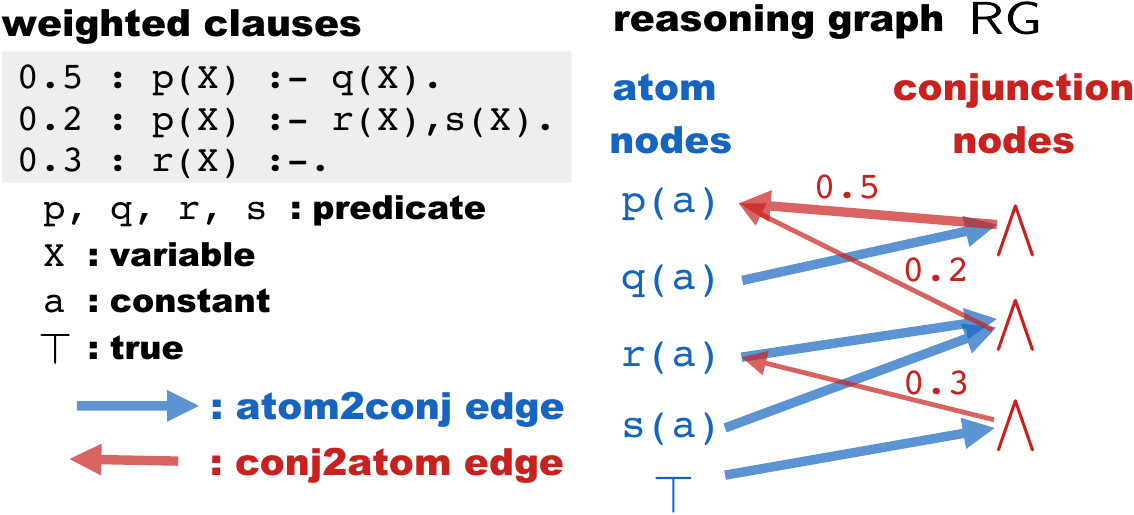}
    \caption{\textbf{Example Forward Reasoning Graph}.
    Weighted clauses (left) and a corresponding reasoning graph (right).
    Blue nodes represent ground atoms, and red nodes represent conjunctions.  
    Each conjunction node corresponds to each ground clause.
    Edges represent how the ground clauses connect the ground atoms and conjunctions. (Best viewed in color) }
    \label{fig:reasoning_graph}
\end{figure}

We represent a set of weighted clauses as a directed bipartite graph.
Fig. \ref{fig:reasoning_graph} shows an example of a set of weighted clauses and a corresponding forward reasoning graph.
Intuitively, the graph has two groups of nodes representing nodes of ground atoms and nodes of conjunctions. 
Edges represent how the ground clauses connect the ground atoms and conjunctions with their weights.

\begin{definition}
A {\it Forward Reasoning Graph} is a bipartite directed graph $(\mathcal{V}_\mathcal{G},$ $\mathcal{V}_\land, \mathcal{E}_{\mathcal{G} \rightarrow \land}, \mathcal{E}_{\land \rightarrow \mathcal{G}})$, where $\mathcal{V}_\mathcal{G}$ is a set of nodes representing ground atoms (atom nodes), $\mathcal{V}_{\land}$ is set of nodes representing conjunctions (conjunction nodes), $\mathcal{E}_{\mathcal{G} \rightarrow \land}$ is set of edges from atom to conjunction nodes and $\mathcal{E}_{\land \rightarrow \mathcal{G}}$ is a set of edges from conjunction to atom nodes.
\end{definition}

\noindent
Given a set of clauses and ground atoms, Algorithm~\ref{algo:reasoning_graph} shows the construction of a corresponding forward reasoning graph.
\textbf{(Line 1)} First, the graph is initialized by adding the atom nodes for ground atoms $\mathcal{G}$ and a special node $\top$, which represents \emph{true}. It is used to represent clauses that have no body atoms, \eg $\mathtt{r(X)\texttt{:-}}$.
\textbf{(Line 2)} The function $\texttt{ground\_clauses}$ takes a set of clauses and a language as input.
In general, an infinite number of ground terms can be considered with functors in FOL. 
Thus we consider a subset of the ground terms by limiting the number of nested functors. 
A set of ground clauses $\mathcal{C}^*$ is obtained by substituting ground terms for variables.
\textbf{(Line 3--8)} For each ground clause, $C^*_i \in \mathcal{C}^*$, corresponding node and edges are added to the reasoning graph, \ie edges from body atoms to a conjunction, and from a conjunction to a head atom.
$\mathsf{X}$ denotes the corresponding node in the graph for a logical formula $X$.
Each ground clause corresponds to a conjunction node in the reasoning graph.

      \begin{algorithm}[t]
        \caption{Building a forward reasoning graph}
        \begin{algorithmic}[1]
\Require  clauses $\mathcal{C}$, ground atoms $\mathcal{G}$, language $\mathcal{L}$
\State $\mathsf{RG} \leftarrow {\tt init}(\mathcal{G})$  \textcolor{OliveGreen}{\# add $\top$ node and atom nodes for $\mathcal{G}$}
\State $\mathcal{C}^* \leftarrow \texttt{ground\_clauses}(\mathcal{C}, \mathcal{L})$  \textcolor{OliveGreen}{\# compute all possible groundings for clauses $\mathcal{C}$}
\For{$C^*_i =  A ~\mbox{:-}~ B_1, \ldots, B_n  \in \mathcal{C}^*$}
        \State add conjunction node $\mathsf{C^*_i}$ to $\mathsf{RG}$
        \For{$B_j \in [B_1, \ldots, B_n]$}
            \State add edge $(\mathsf{B_j}, \mathsf{C^*_i})$ to $\mathsf{RG}$ \textcolor{OliveGreen}{\# Atom2Conj edge}
        \EndFor
        \State add edge $(\mathsf{C^*_i}, \mathsf{A})$ to $\mathsf{RG}$ \textcolor{OliveGreen}{\# Conj2Atom edge}
        \EndFor
\Ensure $\mathsf{RG}$        
\end{algorithmic}
\label{algo:reasoning_graph}
      \end{algorithm}

\subsection{Message Passing for Forward Chaining}
NEUMANN performs forward-chaining reasoning by passing messages on the reasoning graph.
Essentially, forward reasoning consists of \emph{two} steps: (1) computing conjunctions of body atoms for each clause and (2) computing disjunctions for head atoms deduced by different clauses. These two steps can be efficiently computed on bi-directional message-passing on the forward reasoning graph.
We now describe each step in detail.

\textbf{(Direction $\rightarrow$) From Atom to Conjunction.}
First, messages are passed to the conjunction nodes from atom nodes.
For conjunction node $\mathsf{v_i} \in \mathcal{V}_\land$, the node features are updated:
\begin{align}
    v_i^{(t+1)} = \bigvee \left( v_i^{(t)}, \bigwedge\nolimits_{j \in \mathcal{N}(i)}  v_j^{(t)} \right),
    \label{eq:mp1}
\end{align}
where $\bigwedge$ is a soft implementation of \emph{conjunction}, and  $\bigvee$ is a soft implementation of \emph{disjunction}. 
Intuitively, probabilistic truth values for bodies of all ground clauses are computed softly by Eq.~\ref{eq:mp1}.

\textbf{(Direction $\leftarrow$) From Conjunction to Atom.}
Following the first message passing, the atom nodes are then updated using the messages from conjunction nodes.
For atom node $\mathsf{v_i} \in \mathcal{V}_\mathcal{G}$, the node features are updated:
\begin{align}
    v_i^{(t+1)} = \bigvee \left( v_i^{(t)}, \bigvee\nolimits_{j \in \mathcal{N}(i)} w_{ji} \cdot v_j^{(t)} \right),
    \label{eq:mp2}
\end{align}
where $w_{ji}$ is a weight of edge $e_{j \rightarrow i}$.
We assume that each clause $C_k \in \mathcal{C}$ has its weight $\theta_k$, and $w_{ji} = \theta_k$ if edge $e_{j \rightarrow i}$ on the reasoning graph is produced by clause $C_k$.
Intuitively, in Eq.~\ref{eq:mp2}, new atoms are deduced by gathering values from different ground clauses and from the previous step.

Performing message-passing by Eq.~\ref{eq:mp1}-\ref{eq:mp2} corresponds to deducing new atoms by Eq.~\ref{eq:T_c} in FOL using probabilistic inputs and weighted clauses.
We used product for conjunction, and \emph{log-sum-exp} function~\cite{Cuturi17} for disjunction:
\begin{align}
    \mathit{softor}^\gamma(x_1, \ldots, x_n) = \gamma \log \sum_{1\leq i \leq n} \exp(x_i / \gamma),
    \label{eq:softor}
\end{align}
where $\gamma > 0$ is a smooth parameter. Eq.~\ref{eq:softor} approximates the maximum value given input $x_1, \ldots, x_n$ in a differentiable manner.

\subsection{Prediction}
The probabilistic logical entailment is computed by the bi-directional message-passing.
Let $\mathbf{x}_\mathit{atoms}^{(0)} \in [0,1]^{|\mathcal{G}|}$ be input node features, which map a ground atom to a scalar value, $\mathsf{RG}$ be the reasoning graph, $\mathbf{w}$ be the clause weights, $\mathcal{B}$ be background knowledge, and $T \in \mathbb{N}$ be the infer step.
For ground atom $G_i \in \mathcal{G}$, NEUMANN computes the probability as follows:
\begin{align}
p(G_i ~|~ \mathbf{x}_\mathit{atoms}^{(0)}, \mathsf{RG}, \mathbf{w}, \mathcal{B}, T) = \mathbf{x}^{(T)}_\mathit{atoms}[i],
\label{eq:neumann_prob}
\end{align}
where $\mathbf{x}_\mathit{atoms}^{(T)} \in [0,1]^{|\mathcal{G}|}$ is the node features of atom nodes after $T$-steps of the bi-directional message-passing.

Algorithm~\ref{algo:reasoning} summarizes the reasoning steps on NEUMANN. \textbf{(Line 1)} Input scene $s$ is converted to probabilistic atoms $\mathbf{x}_\mathit{atoms}^{(0)}$ by the perception function $f_\mathit{perceive}$. We used the perception module of $\alpha$ILP~\cite{shindo23alphailp}, which performs visual perception to produce object-centric representations, and converts them to probabilistic atoms.
Given background knowledge is also incorporated to produce $\mathbf{x}^{(0)}_\mathit{atoms}$.
\textbf{(Line 2-4)} For each reasoning time step, the messages are propagated from the atom nodes to the conjunction nodes by Eq.~\ref{eq:mp1}. %
\textbf{(Line 5-6)} The messages are propagated from the conjunction nodes to the atom nodes by Eq.~\ref{eq:mp2}.
\textbf{(Line 8-9)} The value for the target atom $G_i$ is extracted by Eq.~\ref{eq:neumann_prob} and returned.
\begin{algorithm}[t]
        \caption{Reasoning on NEUMANN}
        \begin{algorithmic}[1]
\Require input scene $s$, reasoning graph $\mathsf{RG}$, clause weights $\mathbf{w}$, background knowledge $\mathcal{B}$, reasoning step $T$, target atom $G_i$
\State $\mathbf{x}_\mathit{atoms}^{(0)} = f_\mathit{perceive}(s, \mathcal{B})$ \textcolor{OliveGreen}{\# visual perception}
\For{$t \in [1, \ldots, T ]$} \\ \hspace{1.3em} \textcolor{OliveGreen}{\# massages from atom nodes to conjunction nodes}
\State $\mathbf{x}_\mathit{conj}^{(t)} = \mathit{atom2conj}(\mathbf{x}_\mathit{atoms}^{(t-1)}, \mathsf{RG})$  \\ \hspace{1.3em} \textcolor{OliveGreen}{\# massages from conjunction nodes to atom nodes using clause weights}
\State $\mathbf{x}_\mathit{atoms}^{(t)} = \mathit{conj2atom}(\mathbf{x}_\mathit{conj}^{(t)}, \mathsf{RG}, \mathbf{w})$ 
\EndFor\\
\textcolor{OliveGreen}{\# extract the value of the target atom $G_i$}
\State $p(G_i ~|~ \mathbf{x}_\mathit{atoms}^{(0)}, \mathsf{RG}, \mathbf{w}, \mathcal{B}, T) = \mathbf{x}^{(T)}_\mathit{atoms}[i]$
\Ensure $p(G_i ~|~ \mathbf{x}_\mathit{atoms}^{(0)}, \mathsf{RG}, \mathbf{w}, \mathcal{B}, T)$
\end{algorithmic}
\label{algo:reasoning}
\end{algorithm}

\subsection{NEUMANN Memory Consumption}
We now compare NEUMANN to conventional differentiable forward reasoners~\cite{Evans18,Shindo21,shindo23alphailp}.
NEUMANN achieves memory-efficient reasoning by message-passing.

\begin{proposition}
Let $\mathcal{G}$ be a set of ground atoms and $\mathcal{C}$ be a set of clauses, which produce a set of ground clauses $\mathcal{C}^*$ with a language $\mathcal{L}$.
The memory consumption of the reasoning graph is $\mathcal{O}\left( |\mathcal{G}| + |\mathcal{C}^*| \right)$, while that of the conventional differentiable forward-chaining tensors is $\mathcal{O}\left( |\mathcal{G}| \times |\mathcal{C}^*|  \right)$.
\label{prop:memory}
\end{proposition}

\textbf{Proof.}
The number of atom nodes is $|\mathcal{G}|$, and the number of the conjunction nodes is $|\mathcal{C}^*|.$ Thus, the memory consumption by the nodes is $\mathcal{O}(|\mathcal{G}| + |\mathcal{C}^*|)$.
For each ground clause $C^* = A ~\text{:-}~ B_1, \ldots, B_n \in \mathcal{C}^*$, 
each body atom $B_i$ is connected to a conjunction node, \ie $n$ edges, and another edge from the conjunction node to a head atom $A$.
Thus, the memory consumption of the edges is $\mathcal{O}(|\mathcal{C}^*| \times (n+1))$.
To this end, the total memory consumption of the sum of those of the nodes and edges, \ie $\mathcal{O}(|\mathcal{G}| + |\mathcal{C}^*| + |\mathcal{C}^*|(n+1)) \approx \mathcal{O}(|\mathcal{G}| + |\mathcal{C}^*|)$.
The tensor-based reasoners build a tensor $\mathbf{I} \in \mathbb{N}^{|\mathcal{G}| \times |\mathcal{C}^*|}$, which holds the indices of ground atoms for each ground clause. Thus the overall memory consumption is $\mathcal{O}(|\mathcal{G}| \times |\mathcal{C}^*|)$.

\subsection{Learning Logic Programs by NEUMANN}
Now we describe how NEUMANN searches logic programs given a visual ILP problem.

\textbf{Problem Statement.}
Let $\mathcal{Q} = (\mathcal{E}^+, \mathcal{E}^-, \mathcal{B}, \mathcal{L}, \mathcal{Z})$ be a visual ILP problem, where $\mathcal{E}^{+}$ is a set of positive examples, $\mathcal{E}^-$ is a set of negative examples, $\mathcal{B}$ is background knowledge, $\mathcal{L}$ is a language, and $\mathcal{Z}$ is a language bias. Each example is given as a visual scene. The task is to find a logic program that can perform classification correctly based on the attributes and relations of objects in the scenes.

\begin{figure}
     \centering
     \includegraphics[width=\linewidth]{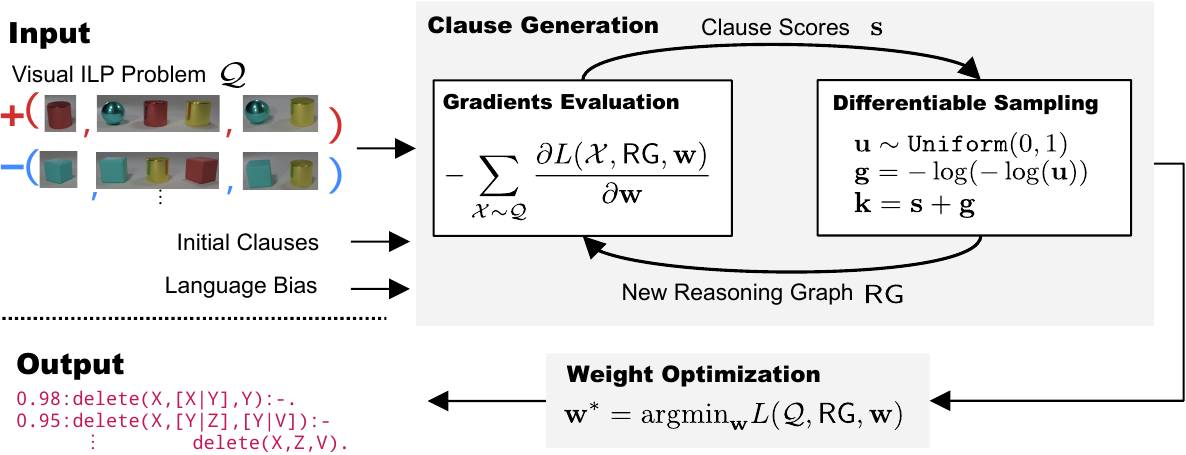}
     \caption{\textbf{Structure learning on NEUMANN}.
     Given positive and negative examples as visual scenes, NEUMANN learns logic programs by the following \emph{two} steps.
(1) NEUMANN generates promising clauses by iterating scoring and sampling of clauses. Candidate clauses are evaluated by computing their gradients for a classification loss, and promising clauses are sampled via differentiable sampling. To this end, new candidate clauses are generated by refining the sampled clauses, and a new reasoning graph is produced.
(2) After the iteration of clause generation steps, NEUMANN assigns randomly-initialized clause weights and optimizes them to minimize the classification loss. 
(Best viewed in color)
}
     \label{fig:learning_neumann}
 \end{figure}

Fig.~\ref{fig:learning_neumann} shows an overview of learning of NEUMANN. It learns logic programs in \emph{two} steps:
(1) NEUMANN generates promising clauses by iterating scoring and sampling of clauses. Candidate clauses are evaluated by computing their gradients for a classification loss, and promising clauses are sampled via differentiable sampling using the Gumbel-max trick. To this end, new candidate clauses are generated by refining the sampled clauses, and a new reasoning graph is produced.
(2) After the iteration of clause generation steps, NEUMANN assigns randomly-initialized clause weights and optimizes them to minimize the classification loss. It uses stochastic gradient descent for the optimization.

We first describe the clause-generation step and weight-optimization step in detail, respectively, and then we explain the whole learning algorithm for NEUMANN, highlighting the difference from existing differentiable ILP solvers.

\subsubsection{Clause Generation}
NEUMANN generates candidate clauses by iteratively (1) scoring clauses using gradients and (2) performing differentiable sampling on the scores and refining them.
We extend the beam search approach used in $\alpha$ILP~\cite{shindo23alphailp} to achieve more efficient clause generation using gradients avoiding nested loops.

\textbf{Clause Scoring by Gradients.}
NEUMANN generates candidates of clauses $\mathcal{C}$ by refining given initial clauses $\mathcal{C}_0$ repeatedly.
We evaluate clauses by computing gradients at once. By using the end-to-end reasoning architecture, NEUMANN scores each clause efficiently using automatic differentiation. Given a visual ILP problem $\mathcal{Q}$, a reasoning graph $\mathsf{RG}$, clause weights $\mathbf{w}$, and background knowledge $\mathcal{B}$, NEUMANN computes the binary-cross entropy loss:
\begin{align}
  L(\mathcal{Q}, \mathsf{RG}, \mathbf{w}) =   -\mathbb{E}_{(e, y) \sim \mathcal{Q}} [ y \log  p(y ~\vert~  &e, \mathsf{RG}, \mathbf{w}, \mathcal{B}, T) ~+ \nonumber \\
    &(1-y) \log (1 -  p(y ~\vert~  e, \mathsf{RG}, \mathbf{w}, \mathcal{B}, T))],
  \label{eq:cross_entropy_loss}
\end{align}
where $(e, y)$ is a tuple of a visual scene $e$ and its label $y$, \ie if $e$ is a positive example then $y = 1$ otherwise $y = 0$.
The conditional probability of the label $p(y ~\vert~  e, \mathsf{RG}, \mathbf{w}, \mathcal{B}, T)$ is computed by using Eq.~\ref{eq:neumann_prob}.

Using the loss, NEUMANN scores candidate clauses $\mathcal{C}$ by computing gradients w.r.t. the clause weights, \ie
NEUMANN computes clause scores $\mathbf{s}\in\mathbb{R}^{|\mathcal{C}|}$:
\begin{align}
  \mathbf{s}  = - \sum_{\mathcal{X} \sim \mathcal{Q}}  \frac{\partial L(\mathcal{X}, \mathsf{RG}, \mathbf{w})}{\partial \mathbf{w}} ,
  \label{eq:grad_score}
 \end{align}
where $\mathcal{X}$ is a sampled batch of labeled examples, $\mathsf{RG}$ is a reasoning graph constructed using clauses $\mathcal{C}$, and $\mathbf{w} \in \mathbb{R}^{|\mathcal{C}|}$ is a clause weight.
Intuitively, useful clauses to classify given visual scenes get negatively large gradients to minimize the classification loss. Thus we compute the negative raw gradients and consider them as the evaluation scores, \ie promising clauses that contribute much to classify examples correctly will get high scores.
For scoring, all clauses are associated with a uniform value to exclude the influence of the difference in weight values. Note that we do not update the clause weights $\mathbf{w}$ in this step but compute gradients to score clauses.

\textbf{Example.} Suppose we want to solve a simple classification of visual scenes with a pattern: \emph{``If there is a red cube, the scene is positive.''}, \eg  a scene in Fig.~\ref{fig:predict} is a positive example. The task is to learn a classification rule in FOL: 
\begin{align*}
    \mathtt{positive(X)\texttt{:-}in(O1,X),color(O1,red),shape(O1,cube).}
\end{align*}
We start from a general clause $\mathtt{positive(X)\texttt{:-}in(O1,X).}$, and by refining, we get, \eg
\begin{align*}
    &\mathtt{positive(X)\texttt{:-}in(O1,X),color(O1,red).}\\
    &\mathtt{positive(X)\texttt{:-}in(O1,X),color(O1,blue).}\\
    &\mathtt{positive(X)\texttt{:-}in(O1,X),color(O1,yellow).}
\end{align*}
We compose a reasoning graph using these three clauses and give a uniform weight to all clauses.
Using the reasoning graph, we compute the scores by Eq.~\ref{eq:grad_score}.
The first clause contributes the most to correct classifications and thus is scored higher than other clauses.
NEUMANN performs inference over given visual scenes only once to score all clauses, not iterating it for each individual clause. %

\begin{figure}[t]
\begin{minipage}{0.5\linewidth}
         \centering
     Scoring in $\alpha$ILP / $\partial$ILP-ST\\
     \includegraphics[width=.99\linewidth]{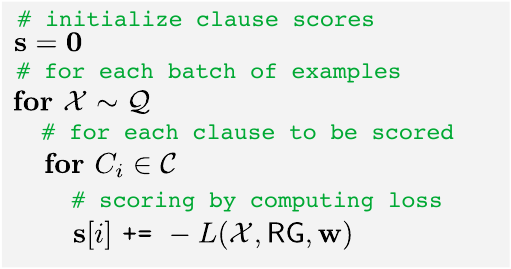}
\end{minipage}
\begin{minipage}{0.5\linewidth}
         \centering
     Scoring in NEUMANN\\
     \includegraphics[width=.99\linewidth]{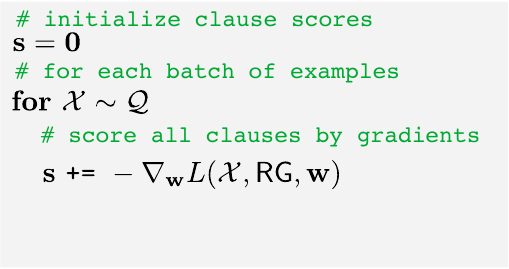}
\end{minipage}
     \caption{\textbf{NEUMANN avoids nested loops for clause scoring.} Input is a visual ILP problem $\mathcal{Q}$ and clauses $\mathcal{C}$, output is the scores $\mathbf{s}$ over $\mathcal{C}$. $\alpha$ILP~\cite{shindo23alphailp} and $\partial$ILP-ST~\cite{Shindo21} evaluate each clause independently.  In contrast, NEUMANN evaluates a set of clauses efficiently without performing for-loop over them.  $\mathcal{X} \sim \mathcal{Q}$ denotes a sampled batch of examples from visual ILP problem $\mathcal{Q}$.}
     \label{fig:clause_eval_diff}
 \end{figure}

Fig.~\ref{fig:clause_eval_diff} illustrates the difference from the conventional clause-scoring strategy. The task is to score candidate clauses $\mathcal{C}$ given visual ILP problem $\mathcal{Q}$ to perform clause search. 
In $\partial$ILP-ST~\cite{Shindo21} and  $\alpha$ILP~\cite{shindo23alphailp}, each clause $C_i \in \mathcal{C}$ needs to be evaluated individually, and thus the computational cost increases quadratically with respect to the number of training data and the number of clauses to be evaluated.
In contrast, NEUMANN evaluates all clauses by calling the backward function once.

\textbf{Generation by Differentiable Sampling.}
Given clause scores $\mathbf{s}$,
we generate new candidates of clauses by performing differentiable sampling based on the Gumbel-max trick~\cite{Jang17gumbelmax,maddison17gumbelmax} and refine them.
The Gumbel-max simulates efficiently sampling procedures given scores in a differentiable manner. For the clause scores $\mathbf{s}$, a noise term is computed as $\mathbf{g} =  -\log( - \log(\mathbf{u}))$ where $\mathbf{u} \sim \mathtt{Uniform}(0,1)$. Then we add the noise to the original scores as $\mathbf{k} = \mathbf{s} + \mathbf{g}$, \ie $\mathbf{k}$ represents scores mixed with a Gumbel noise.
Then clause $C_i \in \mathcal{C}$ is sampled with $i = \mathit{argmax} (k_1, \ldots, k_{|\mathcal{C}|})$.
The sampled clauses are refined using \emph{downward refinement operator}~\cite{Nienhuys97}, which specifies given clauses, \ie generates more specific clauses than the given clause in terms of the number of atoms to be entailed with it.
Given clause $C$ (\eg $\mathtt{p(X,Y)\texttt{:-}.}$), the refinement operator consists of the following \emph{four} specifications: 
(i) add an atom to the body of $C$ (\eg $\mathtt{p(X,Y)\texttt{:-}q(X,Y).}$), 
(ii) substitute a constant to variable in $C$ (\eg $\mathtt{p(X,a)\texttt{:-}.}$), 
(iii)  remove a variable by substituting another variable in $C$ 
 ($\mathtt{p(X,X)\texttt{:-}.}$), and 
(iv) apply a functor (\eg $\mathtt{p(X,f(Y,Z))\texttt{:-}.}$). Downward refinement operator ensures \emph{completness}, \ie any clauses that consist of a finite set of symbols can be generated by applying the operator for finite times to the most general clause~\cite{Nienhuys97}.
NEUMANN uses mode declarations~\cite{Muggleton95} (\cf App.~\ref{sec:definition_mode_declarations}) to restrict the search space, and clauses that do not satisfy the declarations will be discarded.
The newly generated clauses are added to the set of clauses $\mathcal{C}$ and evaluate the added clauses by Eq.~\ref{eq:grad_score} in the next step.

\subsubsection{Weight Optimization}
After the clause generation, NEUMANN performs loss minimization for classification with respect to clause weights.
So far, we assumed that we have one clause weight vector. By using softmax, NEUMANN can learn to select one clause out of multiple generated clauses. However, in practice, we should be able to learn logic programs consisting of multiple clauses.

NEUMANN composes differentiable logic programs that consist of multiple clauses as follows:
(1) We fix the target programs' size as $M$, \ie where we try to find a logic program with $M$ clauses out of generated clauses $\mathcal{C}$.
(2) We introduce randomly-initialized $|\mathcal{C}|$-dimensional weights $\mathbf{W} = [ {\bf w}^{(1)}, \ldots, {\bf w}^{(M)} ]$ $(\mathbf{w}^{(j)} \in \mathbb{R}^{|\mathcal{C}|})$, \ie each clause gets $M$ individual weights.
(3) We take softmax of each weight vector ${\bf w}^{(j)} \in \mathbf{W}$ and softly choose $M$ clauses out of $|\mathcal{C}|$ clauses, \ie $\hat{\mathbf{w}}^{(j)} = \mathit{softmax}(w_0^{(j)}, \ldots, w_{|\mathcal{C}|}^{(j)})$.
(4) We compose a clause weight vector $\mathbf{w} \in [0,1]^{|\mathcal{C}|} $ as:
\begin{align}
    w_i = \mathit{softor}^\gamma \left(\hat{w}_i^{(1)}, \ldots, \hat{w}_i^{(M)} \right) =  \gamma \log \sum_{1\leq j \leq M} \exp(\hat{w}_i^{(j)} / \gamma),    
    \label{eq:softor_weight}
\end{align}
where $\gamma>0$ is a smooth parameter, approximating the maximum value out of $M$ weights for each clause in a differentiable manner.

For example, suppose $3$ clauses are generated by NEUMANN, and we want to compose a logic program that consists of $2$ clauses, \ie $|\mathcal{C}| = 3$ and $M = 2$.
By initializing $2$ weight vectors and applying softmax to each, we get, \eg  $\hat{\mathbf{w}}^{(1)} = [0.1, 0.7, 0.2]^\top$ and $\hat{\mathbf{w}}^{(2)} = [0.8, 0.1, 0.1]^\top$. Using Eq.~\ref{eq:softor_weight}, we get $\mathbf{w} \approx [0.8, 0.7, 0.2]^\top$, where the first 2 clauses get large weights.

The weights for the clauses are trained to minimize the loss function. By using the end-to-end reasoning architecture, NEUMANN finds a logic program that explains the complex visual scenes by gradient descent, \ie solves  
\begin{align}
    \mathbf{w}^* = \text{argmin}_\mathbf{w} L(\mathcal{Q}, \mathsf{RG}, \mathbf{w}),
\end{align} where $L$ is the cross-entropy loss (Eq.~\ref{eq:cross_entropy_loss}).
NEUMANN minimizes the loss based on stochastic gradient descent. 
After performing sufficient weight-update steps, the generated clauses and their trained weights are returned.

Algorithm~\ref{algo:learning_overview_algo} shows the entire learning process of NEUMANN.
\textbf{(Line 1-3)} An initial reasoning graph is built. \textbf{(Line 5-10)} Clauses $\mathcal{C}$ are scored by computing gradients. Useful clauses in $\mathcal{C}$ get negatively large gradients, and thus they are scored high at line 10.
\textbf{(Line 13-21)} Sample clauses to be refined to generate new clauses according to the scores using the Gumbel-max trick.
\textbf{(Line 22-25)} The sampled clauses are refined to generate clauses to be scored in the next iteration.
\textbf{(Line 27-32)} NEUMANN performs weight optimization using the generated clauses $\mathcal{C}_\mathit{sampled}$ with randomly initialized clause weights $\mathbf{w}$.

\begin{algorithm}[t]
        \caption{Learning NEUMANN}
        \begin{algorithmic}[1]
\Require visual ILP problem $\mathcal{Q}$, ground atoms $\mathcal{G}$, language $\mathcal{L}$, initial clauses $\mathcal{C}_0$, language bias $\mathcal{Z}$, search parameters $N_\mathit{trial}$, $N_\mathit{samlpe}$, target program size $M$
\State $\mathsf{RG} = {\tt build\_reasoning\_graph}(\mathcal{C}_0, \mathcal{G}, \mathcal{L})$  \textcolor{OliveGreen}{\# initialize a reasoning graph}
\State $\mathcal{C}_\mathit{sampled} = \phi$ \textcolor{OliveGreen}{\# all sampled clauses} 
\State $\mathcal{C} = \mathcal{C}_0$ \textcolor{OliveGreen}{\# clauses to be scored next} \\\textcolor{OliveGreen}{\# perform clause-generation for $N_\mathit{trial}$ times}
\For{$n \in [1, \ldots, N_\mathit{trial} ]$} \\ \hspace{1.3em} \textcolor{Orange}{\# clause evaluation by computing gradients}
\State $\mathbf{s} = \mathbf{0}$ \textcolor{OliveGreen}{\# initialize clause scores} 

\For{$ \mathcal{X} \sim \mathcal{Q}$} \\ \hspace{2.6em} \textcolor{OliveGreen}{\# compute scores by gradients with coefficient $\beta > 0$}
    \State $\mathbf{s} = \mathbf{s}  +  \beta \cdot (- \nabla_\mathbf{w} L(\mathcal{X}, \mathsf{RG}, \mathbf{w}))$ 
    \EndFor
    \\ \hspace{1.3em} \textcolor{Orange}{\# sample $N_\mathit{sample}$ clauses based on the scores}
    \State $\mathcal{D}_\mathit{sampled} = \phi$  \textcolor{OliveGreen}{\# sampled clauses at step $n$}

    \For{$m \in [1, \ldots, N_\mathit{sample}]$} \\ \hspace{2.6em} \textcolor{OliveGreen}{\# sample $N_\mathit{sample}$ clauses using the Gumbel-max trick}
        \State $\mathbf{u} \sim \mathtt{Uniform}(0,1)$
        \State $\mathbf{g} =  -\log( - \log(\mathbf{u}))$
        \State $\mathbf{k} = \mathbf{s} + \mathbf{g}$
        \State $C_i \in \mathcal{C}$ is sampled with $i = \mathit{argmax} (k_1, \ldots, k_{|\mathcal{C}|})$
        \State add $C_i$ to $\mathcal{D}_\mathit{sampled}$
\EndFor \\ \hspace{1.3em}  \textcolor{OliveGreen}{\# generate new clauses to be scored in the next iteration using language bias}
\State $\mathcal{C} = \mathit{downward\_refinement}(\mathcal{D}_\mathit{sampled}, \mathcal{L}, \mathcal{Z})$\\ \hspace{1.3em}
\textcolor{OliveGreen}{\# update all sampled clauses}
\State $\mathcal{C}_\mathit{sampled} = \mathcal{C}_\mathit{sampled} \cup \mathcal{D}_\mathit{sampled}$ 
\EndFor
\\\textcolor{OliveGreen}{\# initialize a reasoning graph using all sampled clauses}
\State $\mathsf{RG} = {\tt build\_reasoning\_graph}(\mathcal{C}_\mathit{sampled}, \mathcal{G}, \mathcal{L})$   \\ \textcolor{OliveGreen}{\# initialize the clause weights according to the target program size $M$}
\State $\mathbf{w} = {\tt initialize\_weights}(\mathcal{C}_\mathit{sampled}, M)$
    \\ \textcolor{Orange}{\# clause weight optimization by stochastic gradient descent}
\State $\mathbf{w}^* = \text{argmin}_\mathbf{w} L(\mathcal{Q}, \mathsf{RG}, \mathbf{w})$
\Ensure $\mathcal{C}_\mathit{sampled}, \mathbf{w}^*$        
\end{algorithmic}
\label{algo:learning_overview_algo}
      \end{algorithm}

We highlight the difference between NEUMANN and other differentiable ILP approaches in terms of the memory cost and clause-search cost in Tab.~\ref{tab:diff_ilp_comparison}.
As shown in Prop.~\ref{prop:memory}, NEUMANN consumes less memory than other approaches, \ie NEUMANN consumes memory linearly with the number of ground atoms and clauses, but others consume quadratically.
$\partial$ILP generates clauses by templates without any symbolic search. Thus it requires no cost for searching but needs to exclude functors to manage the number of clauses to be generated.
$\partial$ILP-ST and $\alpha$ILP perform beam search using exact scoring of clauses.
As illustrated in Fig.~\ref{fig:clause_eval_diff}, the time complexity of exact scoring is $\mathcal{O}(N_\mathit{data} \times |\mathcal{C}| \times R)$, where $N_\mathit{data}$ is the number of data, $\mathcal{C}$ is the set of clauses to be scored, and $R$ is the time complexity of the reasoning function. Although they require nested loops for data and clauses, they can handle functors because beam search can prune redundant clauses.
In contrast, NEUMANN computes forward and backward pass for each data to evaluate clauses $\mathcal{C}$, and thus the time complexity of the scoring is $\mathcal{O}(N_\mathit{data} \times (R + R)) \approx \mathcal{O}(N_\mathit{data} \times R)$ because both forward and backward pass have the time complexity of $R$. The scoring of clauses needs to be conducted at every step of the search, and thus it is crucial to have an efficient scoring strategy.

 \begin{table}[t]
 \caption{\textbf{NEUMANN is a memory-efficient differentiable ILP solver equipped with an efficient learning algorithm.} A comparison of memory consumption, search cost for each step, scoring method, and the capability of handling functors with other differentiable ILP solvers. $\mathcal{G}$ is a set of ground atoms, $\mathcal{C}$ is a set of clauses, and $\mathcal{C}^*$ is a set of ground clauses. $N_\mathit{data}$ is the number of examples, $R$ is the time complexity of the differentiable forward chaining. }
 
\begin{tabular}{ccccc}
 & \begin{tabular}[c]{@{}c@{}}Memory\\ Cost\end{tabular}  
 & \begin{tabular}[c]{@{}c@{}}Search Cost\\ per Step\end{tabular} 
 & \begin{tabular}[c]{@{}c@{}}Scoring\\ Method\end{tabular} 
 & Functors \\ \hline
$\partial$ILP~\cite{Evans18}          
  & $\mathcal{O}(|\mathcal{G}| \times |\mathcal{C}^*|)$    
  & $\mathcal{O}(1)$      
  & No Scoring (Template)    
  & \XMARK     \\
$\partial$ILP-ST~\cite{Shindo21}      
  & $\mathcal{O}(|\mathcal{G}| \times |\mathcal{C}^*|)$  
  & $\mathcal{O}(N_\mathit{data} \times |\mathcal{C}| \times R)$ 
  & Exact Scoring 
  & \CMARK \\
$\alpha$ILP~\cite{shindo23alphailp}   
  & $\mathcal{O}(|\mathcal{G}| \times |\mathcal{C}^*|)$  
  & $\mathcal{O}(N_\mathit{data} \times |\mathcal{C}| \times R)$ 
  & Exact Scoring 
  & \CMARK \\
NEUMANN        
  & $\mathcal{O}(|\mathcal{G}| + |\mathcal{C}^*|)$  
  & $\mathcal{O}(N_\mathit{data} \times R)$ 
  & Gradient-based 
  & \CMARK  \\ \hline
\end{tabular}
\label{tab:diff_ilp_comparison}
\end{table}

\section{Experiments}
We empirically show that NEUMANN is a memory-efficient differentiable forward reasoner equipped with a computationally-efficient learning algorithm by solving visual reasoning tasks. 
Moreover, we show that NEUMANN solves the proposed \emph{Behind-the-Scenes} task, where different model-building abilities are required beyond perception.
To this end, we show that NEUMANN can perform scalable visual reasoning and learning and provide visual explanations efficiently, outperforming existing symbolic and neuro-symbolic benchmarks.
We implemented NEUMANN using PyTorch. All experiments were performed on one NVIDIA A100-SXM4-40GB GPU with Xeon(R):8174 CPU@3.10GHz and 100 GB of RAM.

We aim to answer the following questions:
\begin{enumerate}
\item[]\textbf{Q1:} Does the message-passing reasoning algorithm simulate the differentiable forward reasoning dealing with uncertainty?
\item[]\textbf{Q2:} Can NEUMANN solve visual ILP problems combined with DNNs outperforming neural baselines and consuming less memory than the other differentiable ILP benchmarks?
\item[]\textbf{Q3:} Does NEUMANN solve the Behind-the-Scenes task outperforming conventional differentiable reasoners providing the model-building abilities (\cf Tab.~\ref{tab:compare_tasks})?
\item[]\textbf{Q4:} Does NEUMANN provide advantages over state-of-the-art symbolic and neuro-symbolic methods? 
\end{enumerate}
\subsection{Differentiable Reasoning with Uncertainty}
To answer \textbf{Q1}, we compare NEUMANN with a conventional tensor-based differentiable forward reasoner, $\alpha$ILP~\cite{shindo23alphailp}, and show that both reasoners produce almost the same proof histories dealing with uncertainties given the same input.
We explore two datasets used in $\partial$ILP~\cite{Evans18}.

\textbf{Even/Odd.} Even/Odd is a synthetic dataset to classify even and odd numbers.
We used the following program:
\begin{align*}
    \mathtt{1.0: even(s^2(X))\texttt{:-}even(X).}
\end{align*}
and background knowledge $\mathcal{B} = \{ \mathtt{even(0)} \}$. $\mathtt{s}$ is a functor that represents \emph{sucessor} of natural numbers, \eg natural number $2$ can be represented by a term $\mathtt{s(s(0))}$.
The task is to deduce even numbers given the rule about even numbers and the base fact that $0$ is an even number.

\begin{figure}
    \centering
    \includegraphics[width=\linewidth]{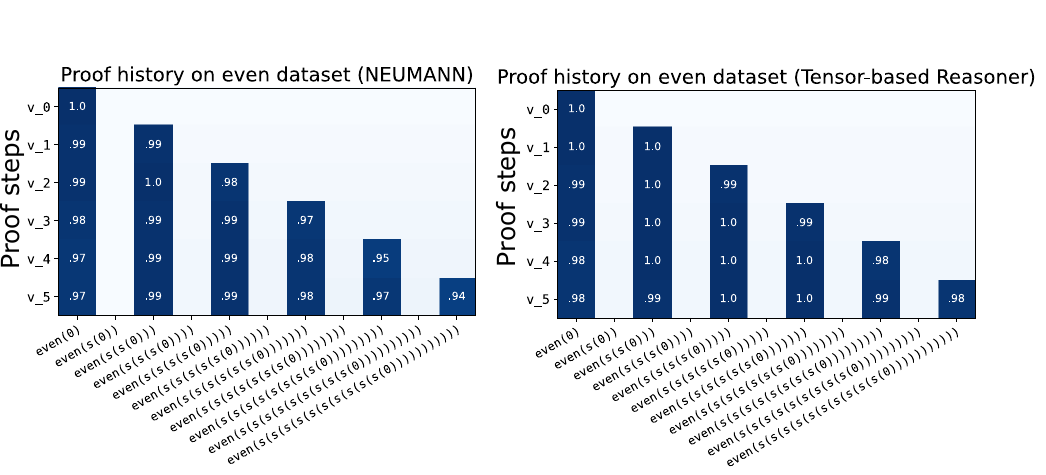}
    \caption{\textbf{NEUMANN performs differentiable forward reasoning.} Proof histories in the Even/Odd task~\cite{Evans18} by NEUMANN (left) and $\alpha$ILP~\cite{shindo23alphailp}. NEUMANN produces almost the same values as the tensor-based reasoner.}
    \label{fig:even_proof}
\end{figure}

Fig.~\ref{fig:even_proof} shows the proof history produced by NEUMANN and $\alpha$ILP for the even/odd dataset.
Each element of the $x$-axis represents a ground atom.
For the $y$-axis, from top to bottom, each row represents a vector of probabilities over atoms for $5$-steps of differentiable forward reasoning, \ie $\mathbf{x}^{(0)}_\mathit{atoms}, \ldots, \mathbf{x}^{(5)}_\mathit{atoms}$.
Both reasoners deduced step by step the following atoms, $\mathtt{even(0), even(s^2(0))},$ $\ldots, \mathtt{even(s^{10}(0))}$, with high probabilities, which are almost $1.0$, but not any odd numbers, \ie they successfully deduced even numbers producing almost the same proof histories.
The message-passing algorithm simulates forward reasoning (Eq.~\ref{eq:T_c}) in FOL. 

\begin{figure}
    \centering
    \includegraphics[width=\linewidth]{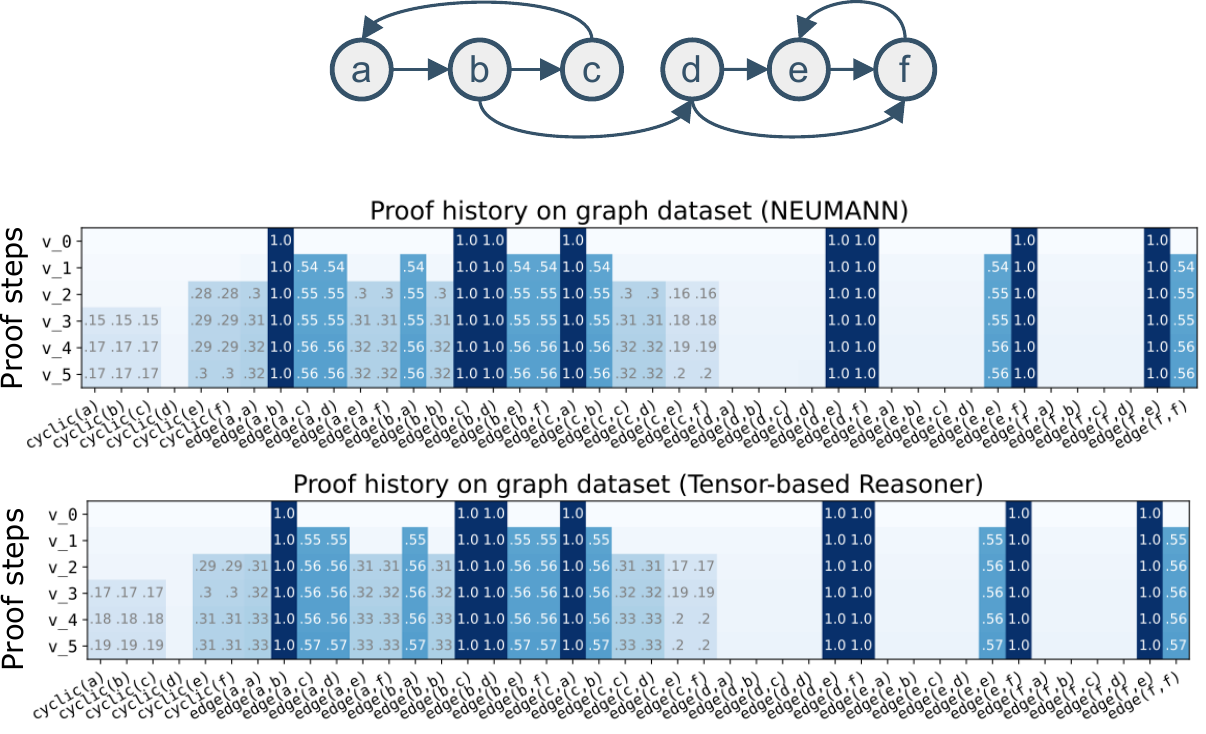}
  \caption{\textbf{NEUMANN performs differentiable forward reasoning.} A directed graph (top) and proof histories in the Cyclic Graph task~\cite{Evans18} by NEUMANN and $\alpha$ILP~\cite{shindo23alphailp} (bottom). NEUMANN produces almost the same values as the tensor-based reasoner dealing with uncertainty.}   
  \label{fig:graph_proof}
\end{figure}

\textbf{Cyclic Graph.}
Cyclic Graph is a dataset to classify if each node in a graph is cyclic or not. We used the same directed graph used in $\partial$ILP~\cite{Evans18} as shown in Fig.~\ref{fig:graph_proof} (top).
We used the following weighted clauses to describe the rules of cyclicity:
\begin{align*}
    &\mathtt{0.51: cyclic(X)\texttt{:-}edge(X,X).}\\
    &\mathtt{0.54: edge(X,Y)\texttt{:-}edge(X,Z),edge(Z,Y).}
\end{align*}
and background knowledge to represent the graph:
\begin{align*}
\mathcal{B} = \left\{
\begin{array}{cccc}
     \mathtt{edge(a,b)}, & \mathtt{edge(b,c)}, & \mathtt{edge(b,d)}, & \mathtt{edge(c,a)}, \\
     \mathtt{edge(d,e)}, & \mathtt{edge(d,f)}, & \mathtt{edge(e,f)}, & \mathtt{edge(f,e)}
\end{array}
\right\}.
\end{align*}
$\mathtt{cyclic(X)}$ means node $\mathtt{X}$ is a cyclic node, \ie there is a path to trace that starts and ends at node $\mathtt{X}$.
The task is to deduce whether each node is a cyclic node or not, given the set of nodes and weighted clauses.

Fig.~\ref{fig:graph_proof} shows a proof history produced by NEUMANN and $\alpha$ILP for the Cyclic Graph dataset.
We show the proof history of the $5$-steps of differentiable forward reasoning. 
Both reasoners produced almost the same probabilities for each ground atom at each time step.
More importantly, both reasoners deal with uncertainty, \ie since the given programs have different weights and thus reasoners need to compute probabilistic values for each ground atom according to the weights, and NEUMANN successfully simulates the differentiable forward reasoning by the message-passing algorithm.
These results show that the message-passing reasoning on NEUMANN is a valid differentiable forward reasoning function, even though it consumes much less memory than the tensor-based differentiable forward reasoners, \eg $\partial$ILP, $\partial$ILP-ST, and $\alpha$ILP.

\subsection{Differentiable ILP on Complex Visual Scenes}
To answer \textbf{Q2}, we compare the performance of NEUMANN with conventional differentiable ILP solvers and neural baselines on visual reasoning tasks and show obtained explanatory programs. 
We also compare the memory consumption of NEUMANN with conventional differentiable forward reasoners.

\begin{figure}
    \centering
    \includegraphics[width=\linewidth]{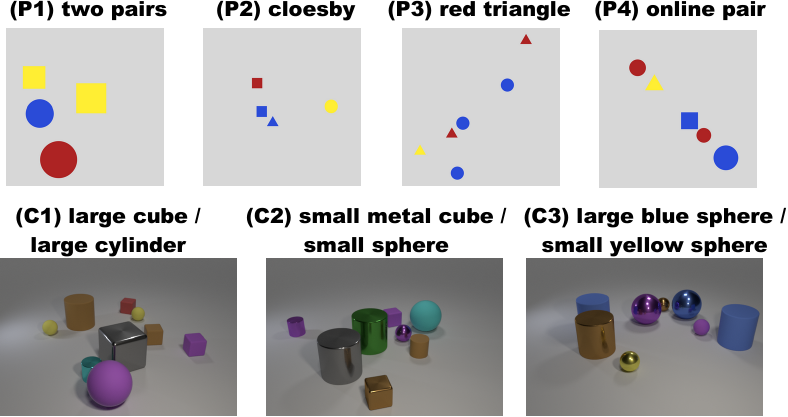}
    \caption{\textbf{Examples of Kandinsky patterns and CLEVR-Hans datasets.} The task is to classify visual scenes by obtaining explicit classification rules. Each Kandinsky pattern is characterized as follows: 
     (\textbf{twopairs}) {\em A pair with the same shape and color, and another pair with the same shape and different colors.} 
(\textbf{closeby})       {\em Two objects closely located.}
 (\textbf{red-triangle}) {\em A red triangle close to a non-triangle and non-red object.}
(\textbf{online-pair}) {\em Five objects aligned on a line and a pair with the same shape and color.} 
(\textbf{long-line}) {\em Seven objects aligned on a line.} 
(Best viewed in color)}
    \label{fig:kandinsky_clevr}
\end{figure}

\textbf{Dataset.}
We adopted Kandinsky patterns~\cite{Muller21kandinsky} and CLEVR-Hans~\cite{Stammer21} dataset.
Both datasets are defined as a classification task of visual scenes, and the classification rules are defined by attributes of the objects and their relations.
Fig.~\ref{fig:kandinsky_clevr} shows examples of the patterns we used.
We use 5 Kandinsky patterns:  \emph{(P1) twopairs}, \emph{(P2) closeby}, \emph{(P3) red-triangle}, \emph{(P4) online-pair}, and \emph{(P5) long-line}. (P5) is an extension of (P4) where the number of objects is increased to $7$ and the constraints of pairing are removed.
We performed structure learning on (P1)-(P4), and for (P5), we used a given clause\footnote{$\mathtt{kp5(X):-in(O1,X),in(O2,X),in(O3,X),in(O4,X),}$ $\mathtt{in(O5,X),in(O6,X),in(O7,X),online(O1,O2,O3,O4,O5,O6,O7).}$} to perform reasoning to assess the scalability of logic reasoners for many objects. 
The dataset contains 10k training examples for each pattern for each positive and negative class, respectively. Likewise, each validation and test split contains 5k examples for each positive and negative class.  
The CLEVR-Hans dataset~\cite{Stammer21} contains CLEVR~\cite{Johnson17} 3D images, and each image is associated with a class label. Examples for each are shown in Fig.~\ref{fig:kandinsky_clevr}. We consider $3$ binary classification tasks for each pattern. Each class contains 3k training images, 750 validation images, and 750 test images, respectively.
More examples for both datasets are available in App.~\ref{sec:more_examples}.

\textbf{Models.}
For Kandinsky patterns, we compare NEUMANN against \emph{two} neural baselines and a differentiable ILP baseline.
We adopted the ResNet-based CNN model~\cite{He16} as a benchmark and also an object-centric benchmark, YOLO+MLP, where the input figure is fed to the pre-trained YOLO model~\cite{Redmon16}. %
The output of the pre-trained YOLO model is fed into MLP with $2$ hidden layers with nonlinearity to predict the class label.
We trained the whole YOLO+MLP network jointly.
For CLEVR-Hans tasks, the considered baselines are the ResNet-based CNN model~\cite{He16}, and the Neuro-Symbolic (NeSy) model~\cite{Stammer21}. The NeSy model uses slot attention~\cite{Locatello20} to perceive objects and feeds its output to Set Transformer~\cite{Juho19}. NeSy-XIL is a NeSy model trained using additional supervision on their explanations. NeSy-XIL is the SOTA neural baseline in the CLEVR-Hans dataset. 
We used $\alpha$ILP~\cite{shindo23alphailp} as a differentiable ILP baseline for both tasks, and we used the mode declarations~\cite{Muggleton95}, a commonly used language bias in ILP.  
The perception networks (YOLO and slot attention) are also used for NEUMANN and $\alpha$ILP in Kandinsky patterns and CLEVR-Hans, respectively.
More details about each baseline are in App.~\ref{sec:detail_kandinsky_clevrhans}.

\textbf{Result.}
We show the accuracy in the test split of the Kandinsky patterns in Tab. \ref{tab:kp}. 
Both $\alpha$ILP and NEUMANN achieved perfect accuracies for each pattern, although CNN's over-fit while training and performed poorly with testing data.
In (P5), the $\alpha$ILP produced \emph{run-out-of-memory}\footnote{It is not trivial to distribute the tensor-based forward reasoners to several GPUs because each of the instances requires the whole \emph{index tensor}, \ie if we split the large index tensor to distribute, each distributed reasoner cannot refer some atoms and clauses, and thus the reasoning result will be incomplete. Tensor parallelism requires non-trivial engineering~\cite{Deepak21GPUs,weng2021largeGPUs}. }
because the pattern involves \emph{seven} objects, and many ground atoms and clauses are produced. NEUMANN can perform scalable visual reasoning beyond tensor-based reasoners.
Moreover, we compare the memory consumption of NEUAMANN and $\alpha$ILP on Kandinsky patterns in Tab. \ref{tab:memory_kandinsky}.
NEUMANN clearly consumes less memory than $\alpha$ILP for each pattern, \eg NEUMANN's reasoning graph size is just $3.12\%$ of the tensor size produced by $\alpha$ILP.
These results show the memory efficiency of the message-passing reasoners of NEUMANN.

Tab. \ref{tab:ch} shows the results for the CLEVR-Hans dataset. As one can see, 
NEUMANN achieved high accuracy similar to $\alpha$ILP, outperforming neural-based baselines, showing the capability of NEUMANN  in complex 3D visual scenes. Moreover, CLEVR-Hans is a confounded dataset, \eg for the first pattern (C1), a large \emph{gray} cube and a large cylinder appear in the training and validation scenes, but in the test scenes, the large cube can be different colors.
Thus, neural baselines perform poorly on the test split because the pure data-driven neural models can be easily confounded~\cite{Stammer21}.
Both $\alpha$ILP and NEUMANN achieved high accuracy in the test split because the \emph{downward refinement}-based clause generation can control the generality of clauses, \ie prevent generating too specific clauses to avoid over-fit. This is achieved by trying the small number of search steps in the clause generation and increasing it step by step by checking the performance in the validation split.

We observed peaked weight distributions after training, \ie only one clause gets a large weight.
The classification rules obtained by discretizing the clause weights by taking \emph{argmax} for Kandinsky patterns and CLEVR-Hans are shown in Fig.~\ref{fig:clauses_kandinsky_clevr}. NEUMANN discovered explanatory clauses for each visual pattern. 
This shows that NEUMANN's learning algorithm can find proper classification rules given positive and negative examples as complex visual scenes.

\begin{table}[t]
\centering
\setlength\tabcolsep{3pt}
\caption{\textbf{Efficiency of NEUMANN does not sacrifice accuracy.} The mean classification accuracy in the test split in the Kandinsky patterns dataset over 5 random seeds. OOM denotes out-of-memory on a single GPU. Both $\alpha$ILP and NEUMANN achieved perfect accuracies for each pattern, although CNN's over-fit while training and performed poorly with testing data. As we mentioned in main text, for pattern P5, we evaluated $\alpha$ILP and NEUMANN with a given program and no structure-learning has been performed. Best results are bold. }
\begin{tabular}{lccccc}
\multicolumn{1}{c|}{Model} & \multicolumn{1}{c|}{(P1) pairs} & \multicolumn{1}{c|}{(P2) close} & \multicolumn{1}{c|}{(P3) red tri.} & \multicolumn{1}{c|}{(P4) online} & \multicolumn{1}{c}{(P5) line} \\ 
\hline
\multicolumn{1}{l|}{CNN}         & $50.0$     & $52.33$    & $55.0$     & \multicolumn{1}{c|}{$50.59$} & - \\
\multicolumn{1}{l|}{YOLO+MLP}    & $99.0$     & $72.93$    & $82.95$    & \multicolumn{1}{c|}{$80.18$} & - \\ \hline
\multicolumn{1}{l|}{$\alpha$ILP} & $\mathbf{100.0}$ & $\mathbf{100.0}$ & $\mathbf{100.0}$ & $\mathbf{100.0}$ & \multicolumn{1}{|c}{\textcolor{red}{OOM}} \\ 
\multicolumn{1}{l|}{NEUMANN}     & $\mathbf{100.0}$ & $\mathbf{100.0}$ & $\mathbf{100.0}$ & $\mathbf{100.0}$ & \multicolumn{1}{|c}{$\mathbf{100.0}$}
\end{tabular}
\label{tab:kp}
\end{table}

\begin{table}[t]
\centering
\scriptsize
\begin{minipage}{.5\linewidth}
\caption{\textbf{NEUMANN is memory efficient and scales beyond tensors.} Memory consumptions on Kandinsky patterns and the visual ILP dataset are shown. The ratio is $\mbox{graph\_size} / \mbox{tensor\_size}$. OOM denotes out-of-memory on a single GPU.}
\begin{tabular}{ccc}
 & \underline{NEUMANN} & \underline{$\alpha$ILP}\\
Dataset      & graph size  (ratio)                                 & tensor size                             \\ \hline
two pairs & $\mathbf{6957}$ ($12.4\%$)  & $56064$                                 \\
closeby    & $\mathbf{1173}$ ($75.8\%$)  & $1548$                                  \\
red-tri. & $\mathbf{4079}$ ($3.12\%$)  & $130410$                                \\
online-p.  &  $\mathbf{15637}$ ($3.13\%$) & $498960$                                \\
long-l.   & $\mathbf{55632}$    (\textcolor{red}{NaN} $\%$)       & \textcolor{red}{OOM} \\ 
\end{tabular}
\label{tab:memory_kandinsky}
\end{minipage}
\begin{minipage}{.4\linewidth}
 \caption{\textbf{Accuracy for CLEVR-Hans dataset} compared to baselines over 5 random seeds. For $\alpha$ILP and NEUMANN, we report the mean over the \emph{three} ILP problems. $\bullet$ denotes best, $\circ$ denotes second best result.}
\begin{tabular}{l|cc}
\multicolumn{1}{c|}{Model} & \multicolumn{1}{c|}{Validation} & Test  \\ \hline
CNN                        & $\circ99.55$                           & $70.34$ \\
NeSy           & $98.55$                           & $81.71$ \\
NeSy-XIL                   & $\bullet100.00$                          & $91.31$ \\ \hline
$\alpha$ILP & $97.5$ & $\bullet \mathbf{97.52}$\\
NEUMANN & $96.67 $ & $\circ\mathbf{97.43}$ 
\end{tabular}
 \label{tab:ch}
\end{minipage}
\label{tab:memory}
\end{table}

\begin{figure}[t]
        \centering
           \begin{lstlisting}[language=Prolog,  style=Prolog-pygsty]
% A pair with the same shape and color, and another
% pair with the same shape and different colors.
kp1(X):-in(O1,X),in(O2,X),in(O3,X),in(O4,X),
        same_shape_pair(O1,O2),same_color_pair(O1,O2),
        same_shape_pair(O3,O4),diff_color_pair(O3,O4).
% Two objects closely located.
kp2(X):-in(O1,X),in(O2,X),closeby(O1,O2).
% A red triangle close to a non-triangle and non-red object.
kp3(X):-in(O1,X),in(O2,X),closeby(O1,O2),
        color(O1,red),shape(O1,triangle),
        diff_shape_pair(O1,O2),diff_color_pair(O1,O2).
% Five objects aligned on a line, and a pair with the same 
% shape and color.
kp4(X):-in(O1,X),in(O2,X),in(O3,X),in(O4,X),in(O5,X),
        same_shape_pair(O1,O2),same_color_pair(O1,O2),
        online(O1,O2,O3,O4,O5).
        
% There is a large cube and a large cylinder.
ch1(X):-in(O1,X),in(O2,X),size(O1,large),shape(O1,cube),
        size(O2,large),shape(O2,cylinder).
% There is a small metal cube and a small sphere.
ch2(X):-in(O1,X),in(O2,X),
        size(O1,small),material(O1,metal),shape(O1,cube),
        size(O2,small),shape(O2,sphere).
% There is a large blue sphere and a small yellow sphere.
ch3(X):-in(O1,X),in(O2,X),size(O1,large),
        color(O1,blue),shape(O1,sphere),
        size(O2,small),color(O2,yellow),shape(O2,sphere).
\end{lstlisting}
    \caption{\textbf{Clauses discovered by NEUMANN for Kandinsky patterns and CLEVR-Hans}. The first 4 clauses for Kandinsky patterns and the last 3 clauses for CLEVR-Hans. }        
    \label{fig:clauses_kandinsky_clevr}
\end{figure}

\begin{figure}[t]
    \centering
    \includegraphics[width=.9\linewidth]{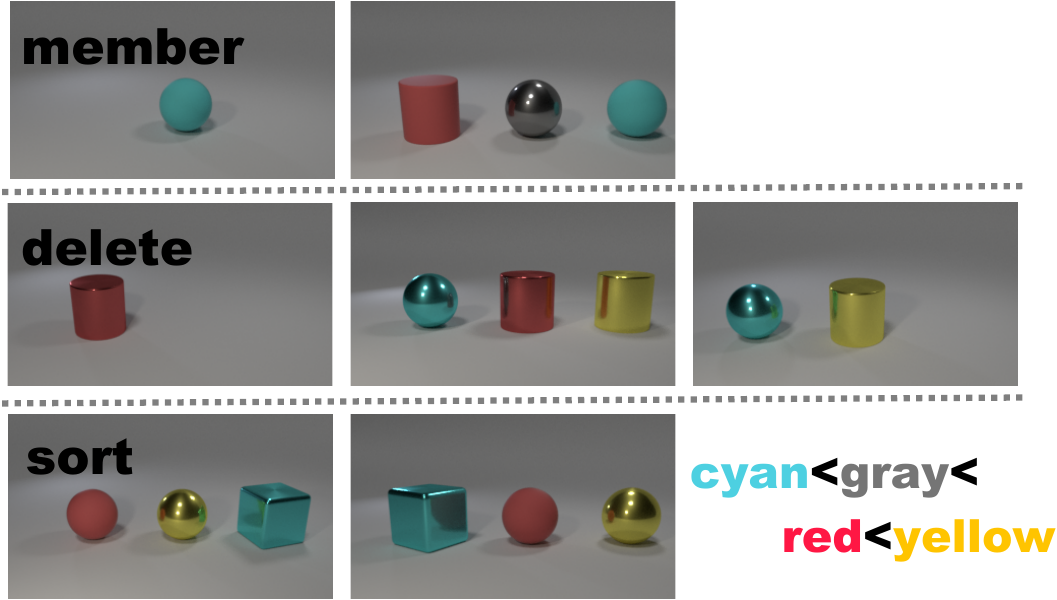}
    \caption{\textbf{Task 1: Abstract Program Induction from Visual Scenes (CLEVR-List).} Positive examples for abstract list operations of \emph{member}, \emph{delete}, and \emph{sort} (with an order of colors: \emph{cyan} $<$ \emph{gray} $<$ \emph{red} $<$ \emph{yellow}, alphabetical order). Each example consists of several visual scenes representing the input and output of the target programs to be learned. The agents need to handle multiple visual scenes, understanding them deely and comparing each other. (Best viewed in color) %
    }
    \label{fig:vilp_examples}
\end{figure}

\begin{figure}[t]
    \centering
    \includegraphics[width=.9\linewidth]{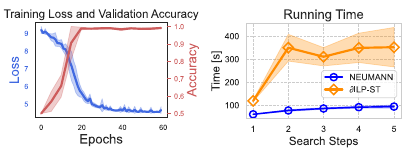}
    \vspace{-1em}
    \caption{\textbf{NEUMANN achieves robust and fast learning.} Training loss and validation accuracy for learning the \emph{delete} operation by NEUMANN (left), and comparison of learning time of logic programs against $\partial$ILP-ST~\cite{Shindo21}, measured for each clause generation step (right). Computed for $5$ different random seeds. (Best viewed in color)}
    \label{fig:delete_loss_acc}
\end{figure}

\begin{figure}[t]
        \centering
           \begin{lstlisting}[language=Prolog,  style=Prolog-pygsty]
% member(X,Y) means X is an element of Y.
member(X,[X|Y]):-.
member(X,[Y|Z]):-member(X,Z).

% delete(X,Y,Z) means Z is the result of deleting X from Y.
delete(X,[X|Y],Y]):-.
delete(X,[Y|Z],[Y|V]):-delete(X,Z,V).

% sort(X,Y) means Y is the result of sorting X.
is_sorted([X,Y|Z]):-smaller(X,Y),is_sorted([Y|Z]).
is_sorted([X]):-.
sort(X,Y):-permutation(X,Y),is_sorted(Y).
\end{lstlisting}
    \caption{\textbf{Abstract operations discovered by NEUMANN for CLEVR-List.} NEUMANN learns abstract list operations from visual scenes dealing with functors.}        
    \label{fig:clauses_list}
\end{figure}

\subsection{Visual Reasoning Behind the Scenes}
To answer \textbf{Q3}, we compare the performance of NEUMANN with conventional differentiable forward reasoners on the behind-the-scenes task showing that (i) it can learn from small data, (ii) it can handle complex visual scenes, (iii) it can learn explanatory programs, and (iv) it can reason about non-observational scenes.
Moreover, we also compare the running time of the clause search with a differentiable ILP benchmark.

The task consists mainly of \emph{two} parts: \textbf{(Task 1)} learning abstract operations, and  \textbf{(Task 2)} solving queries with \emph{imagination}, as shown in Fig.~\ref{fig:behind-the-scene}. We describe each task in detail.

\subsubsection{Task 1: Learning Abstract Operations - CLEVR-List}
The first task is inductive logic programming from visual scenes for list operations: the \emph{member}, \emph{delete}, and \emph{sort} functions.
This is a 3D visual realization of ILP tasks with structured examples, where the goal is to learn abstract list operations given observed input-output pairs, which has been a long-standing task in classical symbolic ILP settings~\cite{Shapiro83,Caferra13}, and addressed in the differentiable ILP setting recently~\cite{Shindo21}. 

We propose \emph{CLEVR-List}, a visual realization of the list-program induction task by using the CLEVR environment~\cite{Johnson17}, which allows users to generate visual scenes that contain multiple objects with different properties, \eg \emph{large red rubber sphere} and \emph{small gray metal cube}.
Fig.~\ref{fig:vilp_examples} shows positive examples for the member, delete and sort functions, which consist of several images representing the inputs and outputs of the target programs to be learned.
For example, the first row in the figure represents a positive example for $\mathtt{member(\clevrobj{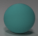}, [~\clevrobj{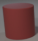},\clevrobj{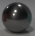},\clevrobj{cyan_sphere.png}~])}$. 
In each training, validation, and test split, we used 200 examples for each positive and negative label, respectively.

Compared to the previously addressed visual reasoning benchmarks, \emph{e.g.} Kandinsky patterns and CLEVR-Hans, CLEVR-List is challenging in the sense that the agents need to handle multiple visual scenes as input, \ie deeply understanding them and comparing each other.
The list programs are involved with functors, and thus, models need to have the capacity to deal with a large number of ground atoms produced by functors.
CLEVR-List requires the following model-building abilities: learning from small data, handling several visual scenes, and learning explanatory programs.

\textbf{Models.}
We compare the performance of NEUMANN with $\partial$ILP-ST~\cite{Shindo21}, which is a state-of-the-art differentiable ILP solver dealing with functors. 
The parameters for the clause generation $(N_\mathit{trial}, N_\mathit{sample})$ are set to $(5, 10)$ for NEUMANN, and we used the same setting for beam search in $\partial$ILP-ST.
We performed $50$ epochs of weight optimization using the RMSProp optimizer with a learning rate of $1e-2$ and infer step $T=5$ for both models.
We used mode declarations~\cite{Muggleton95} as a language bias.%
The number of nested functors is at most $3$, discarding lists with duplicated elements.
More details including used mode declarations are in App.~\ref{sec:detail_behind-the-scenes}.

\textbf{Result.}
Fig.~\ref{fig:delete_loss_acc} (left) shows the training loss and validation accuracy of NEUMANN in the \emph{delete} task, showing the progress of the classification-loss minimization by gradient descent.
For five different random seeds, 
NEUMANN achieved stable learning by producing small training loss and high validation accuracy.

Fig.~\ref{fig:delete_loss_acc} (right) compares the running time of structure learning of NEUMANN and $\partial$ILP-ST. We measured the running time of each clause generation step.
NEUMANN achieved faster learning using gradient-based scoring and differentiable sampling. 
As highlighted in Tab.~\ref{tab:diff_ilp_comparison}, $\partial$ILP-ST performs exact scoring for each clause.
As the search gets deeper, a large number of clauses tend to be generated because of the large number of combinations of symbols. Thus, the running time of $\partial$ILP-ST increases drastically in the search, but NEUMANN consistently achieved fast structure learning.

We observed peaked weight distributions after training, \ie only some clauses with large weights. 
Fig.~\ref{fig:clauses_list} shows the logic programs learned by NEUMANN, obtained by discretizing clause weights using \emph{argmax} after training. NEUMANN produced explanatory programs that achieved $1.0$ of accuracy in the test split for each operation.
These results show that NEUMANN solved the proposed \textbf{(T1)} CLEVR-List task, outperforming a differentiable ILP baseline by the running time.

\subsubsection{Task 2: Reasoning on Behind-the-Scenes}
The second task is to perform visual reasoning given queries where the answers are derived by the \emph{reasoning behind the scenes}, \eg the agent needs to think of the non-observational scenes with \emph{imagination}.
We consider 4 abstract operations: \emph{delete}, \emph{append}, \emph{reverse}, and \emph{sort}.
The input is a pair of an image and a query represented as an atom. 
For example, a query \emph{``What is the color of the second left-most object after deleting a gray object?''} can be represented as a query atom: $\mathtt{query(q\_delete,gray,2nd)}$, where $\mathtt{q\_delete}$ is a constant that represents the query type about the deletion.
Forward reasoners can solve this task by combining clauses to parse input visual scenes and derive answers.
We used 40k questions (10k for each operation) associated with 10k visual scenes.
We compare the performance of NEUMANN and $\alpha$ILP~\cite{shindo23alphailp}.

\textbf{Image Generation.}
We generated visual scenes using the CLEVR environment~\cite{Johnson17}. 
Each visual scene contains at most $3$ objects with different attributes: (i) colors of \emph{cyan, gray, red,} and \emph{yellow}, (ii) shapes of \emph{sphere, cube,} and \emph{cylinder}, (iii) materials of \emph{metal} and \emph{matte}. We excluded color duplications in a single image.

\begin{figure}[t]
    \centering
    \includegraphics[width=.9\linewidth]{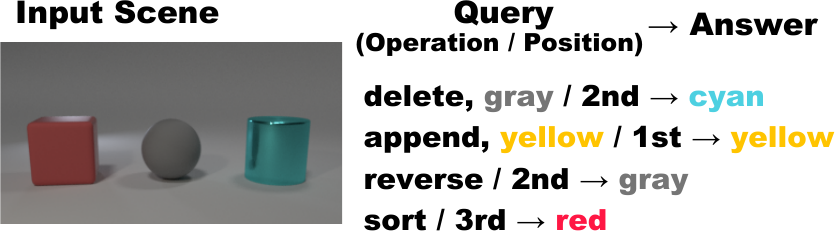}
    \caption{\textbf{Task 2: Visual Reasoning Behind the Scenes.} The task is to compute the answers for given queries paired with input visual scenes.
    A query consists of an \emph{operation} and a target \emph{position}, and an answer is a color, \eg a query \emph{``What is the color of the 2nd left-most object after deleting a gray object?''} whose answer is \emph{cyan}.  (Best viewed in color)} %
    \label{fig:query_answer}
\end{figure}

\textbf{Query Generation.}
We generated queries for the dataset using query templates, which produce various queries using different features of objects.
We used the following template:
\emph{``What is the color of the [Position] object after [Operation]?''},
where [Position] can take either of: \emph{left-most first}, \emph{second}, or \emph{third}.
[Operation] can take the following form: (i) \emph{delete} an object, (ii) \emph{append} an object to the left, (iii) \emph{reverse} the objects, and (iv) \emph{sort} the objects with an order of colors:  %
\emph{cyan} $<$ \emph{gray} $<$ \emph{red} $<$ \emph{yellow}
(alphabetical order). 
Fig.~\ref{fig:query_answer} shows examples of an input scene and some paired queries with their answers.
More examples of input scenes, queries and their answers are in App.~\ref{sec:more_examples}.

\textbf{Models.}
We used the clauses in Fig.~\ref{fig:clauses_behind_the_scenes} for NEUMANN and $\alpha$ILP. 
The first clause about $\mathtt{chain}$ generates a chain of colors given a visual scene.
For example, given the visual scene in Fig.~\ref{fig:query_answer}, the atom $\mathtt{chain([~\clevrobj{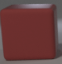},\clevrobj{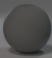},\clevrobj{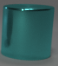}}~])$ is deduced using body atoms $\mathtt{left\_of(\clevrobj{rmc.png},\clevrobj{gms.png})}$ and $\mathtt{left\_of(\clevrobj{gms.png},\clevrobj{cmc.png})}$.
The second clause parses the chained objects to colors. For example, the atom $\mathtt{scene([red,}$ $\mathtt{gray,cyan])}$ is deduced using body atoms of $\mathtt{chain([~\clevrobj{rmc.png},\clevrobj{gms.png},\clevrobj{cmc.png}}~])$, $\mathtt{color(\clevrobj{rmc.png},red)}$, $\mathtt{color(\clevrobj{gms.png},gray)}$, and $\mathtt{color(\clevrobj{cmc.png},cyan)}$.  
The last 4 clauses compute answers for different types of queries using the parsed $\mathtt{scene}$ atoms, list operations, and other utility predicates (\cf Fig.~\ref{fig:clauses_utility} in the appendix ). 
$\mathtt{query2}$ and $\mathtt{query3}$ represent queries, \eg $\mathtt{query2(q, q\_sort, 2nd)}$ represents a query \emph{``What is the color of the 2nd-left object after sorting the objects?''}.
A query atom is given being paired with an input image. %

\begin{figure}[t]
        \centering
           \begin{lstlisting}[language=Prolog,  style=Prolog-pygsty]
% generate a chain of objects from an image
chain([Object1,Object2,Object3]):-
                left_of(Object1,Object2),left_of(Object2,Object3).

% parse the visual scene as a list of colors
scene([Color1,Color2,Color3]):-chain([Object1,Object2,Object3]),
                color(Object1,Color1),
                color(Object2,Color2),
                color(Object3,Color3).

% answer the delete query
answer(X):-scene(Colors1),delete(Color,Colors1,Colors2),
    query3(q_delete,Color,Position),get_color(Colors2,Position,X).
    
% answer the append query
answer(X):-scene(Colors1),append([Color],Colors1,Colors2),
    query3(q_append,Color,Position),get_color(Colors2,Position,X).

% answer the reverse query
answer(X):-scene(Colors1),reverse(Colors1,Colors,Colors2),
    query2(q_reverse,Position),get_color(Colors2,Position,X).

% answer the sort query
answer(X):-scene(Colors1),sort(Colors1,Colors2),
    query2(q_sort,Position),get_color(Colors2,Position,X).
\end{lstlisting}
    \caption{\textbf{Clauses to answer queries for behind-the-scenes.} The first clause about $\mathtt{chain}$ generates a chain of objects given a visual scene.
The second clause parses the chained objects to colors.
The last 4 clauses compute answers for different types of queries using the parsed $\mathtt{scene}$ atoms, list operations, and other utility predicates.}        
    \label{fig:clauses_behind_the_scenes}
\end{figure}

\textbf{Result.}
Tab.~\ref{tab:acc_behind-the-scenes} shows the accuracy for each type of queries.
$\alpha$ILP produced \emph{run-out-of-memory}, however, NEUMANN successfully solves different types of queries with high accuracy.
As shown in Fig.~\ref{fig:clauses_behind_the_scenes}, the clauses to answer the queries consist of many functors for the list representation and existentially quantified variables, and thus a large number of ground atoms and clauses is generated, which is difficult to be handled by the conventional tensor-based reasoner, \ie $\partial$ILP, $\partial$ILP-ST, and $\alpha$ILP.
In fact, Tab.~\ref{table:ground_atoms} shows the numbers of ground atoms and clauses generated for Kandinsky patterns, CLEVR-Hans, and Behind-the-Scenes, respectively.
Behind-the-Scenes requires the models to handle many ground representations, \ie a large \emph{Herbrand base} with functors, and thus memory-efficient reasoning is necessary. 
These results show that NEUMANN solved the Behind-the-Scenes task outperforming conventional tensor-based reasoners overcoming the bottleneck of the intensive memory consumption scaling to deal with functors on visual scenes and query answering.


\begin{table}[t]
\scriptsize
\setlength\tabcolsep{2pt}
\begin{minipage}{.4\linewidth}
\caption{ \textbf{NEUMANN can reason behind the scenes beyond tensor-based  reasoners}. Classification accuracy for behind-the-scenes tasks. OOM denotes out-of-memory on a single GPU.}
\begin{tabular}{c|cccc}
                 & delete                     & append                     & reverse                    & sort                       \\ \hline
$\alpha$ILP & \textcolor{red}{OOM} & \textcolor{red}{OOM} & \textcolor{red}{OOM} & \textcolor{red}{OOM} \\
NEUM.          & $\mathbf{0.98}$                    &       $\mathbf{0.99}$                     &  $\mathbf{0.98}$                          &          $\mathbf{0.98}$
\end{tabular}
 \label{tab:acc_behind-the-scenes}
\end{minipage}
\begin{minipage}{.5\linewidth}
\centering
\setlength\tabcolsep{2pt}
    \caption{\textbf{Many ground representations are required for reasoning about behind-the-scenes.} Number of ground atoms $|\mathcal{G}|$ and ground clauses $|\mathcal{C}^*|$ for each task. For Kandinsky patterns, the mean value for different patterns (P1)-(P4) is shown.}
    \begin{tabular}{c|ccc}
         &Kandinsky & CLEVR H. & Behind-the-Scenes \\\hline
         \#Atoms &$131$ &  $165$ & $150$K \\
         \#Clauses & $288$ & $90$ & $1.1$M
    \end{tabular}
    \label{table:ground_atoms}
\end{minipage}
\end{table}

\subsection{Advantages against other Symbolic and Neuro-Symbolic Methods}
\label{sec:comparison_to_symbolic}
To answer \textbf{Q4}, we compare the performance of NEUMANN against state-of-the-art symbolic and neuro-symbolic methods. Moreover, we show that NEUMANN can produce visual explanations efficiently using gradients using the end-to-end differentiable reasoning architecture.

\subsubsection{Scalable Visual Reasoning and Learning}
\label{sec:inference_time_comparison}
First, we show that NEUMANN can perform scalable visual reasoning, \ie it can handle a large number of examples of complex visual scenes. To show that, 
we compare the inference time for Kandinsky patterns and  CLEVR-Hans.
We used \emph{two} state-of-the-art neuro-symbolic methods to be compared:
\begin{itemize}
    \item \textbf{Feed-Forward Neural-Symbolic Learner (FFNSL)~\cite{CunningtonLLR23ffnsl}}  is a neuro-symbolic learning framework that integrates Answer Set Programming (ASP) with neural networks. It performs visual perception using neural networks, reads out their output as weighted logic representations, and performs Inductive Learning of Answer Set Programs (ILASP)~\cite{ILASP}, which conducts efficient structure-learning of logic programs based on ASP semantics. It uses CLINGO~\cite{gebser_kaminski_kaufmann_schaub_2019clingo}, a well-established ASP-solving system for inference. FFNSL can handle noisy input as weighted expressions, but its inference engine requires discrete input and returns discrete logic representations. %
    \item \textbf{DeepProbLog~\cite{Manhaeve19}} is a neuro-symbolic framework that integrates neural networks to ProbLog~\cite{de2015probabilistic}, which is a well-established probabilistic logic inference engine. ProbLog accepts probabilistic input and produces probabilistic output by performing exact probabilistic inference by compiling logic representations into circuits, \eg Sentential Decision Diagrams~\cite{Darwiche11SDD}. DeepProbLog obtains gradients out of the ProbLog output and train neural networks efficiently, and it is also applied to perform structure learning in the  \emph{sketching} setting~\cite{Solar08sketcing,pbosnjak17dforth}, where programs are learned to complete partially-given input programs.
\end{itemize}
We measured the inference time (including the grounding of programs) for the training split. We changed the proportion of the data to be used from $0.2$ (use $20\%$ of the training data) to $1.0$ (use the full training data) for each dataset.
We used a batch size of 200 consistently throughout the experiments. We set $1000$ seconds as the timeout.

\begin{figure}[t]
\centering
\begin{tabular}{ccc}
\includegraphics[width=0.31\textwidth]{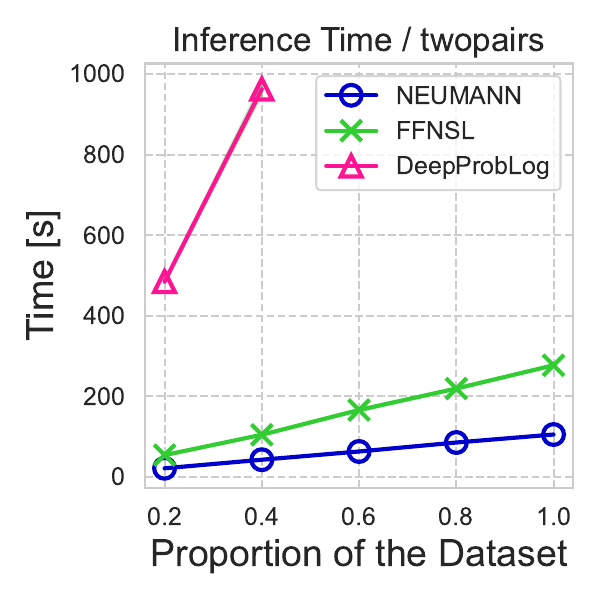} &
\includegraphics[width=0.31\textwidth]{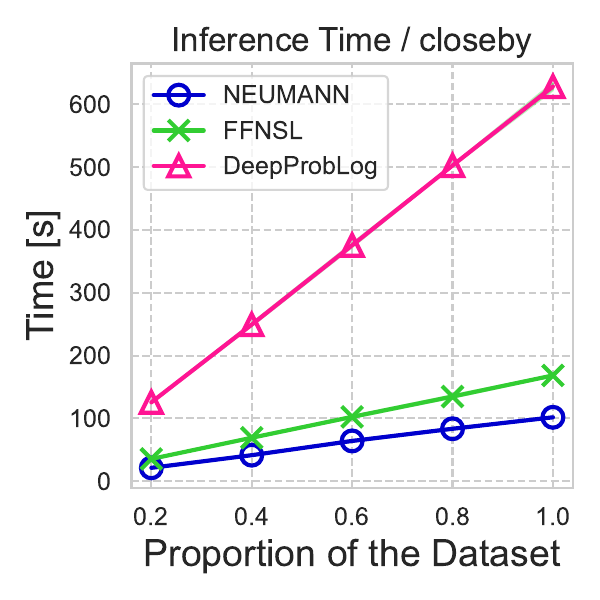} &
\includegraphics[width=0.31\textwidth]{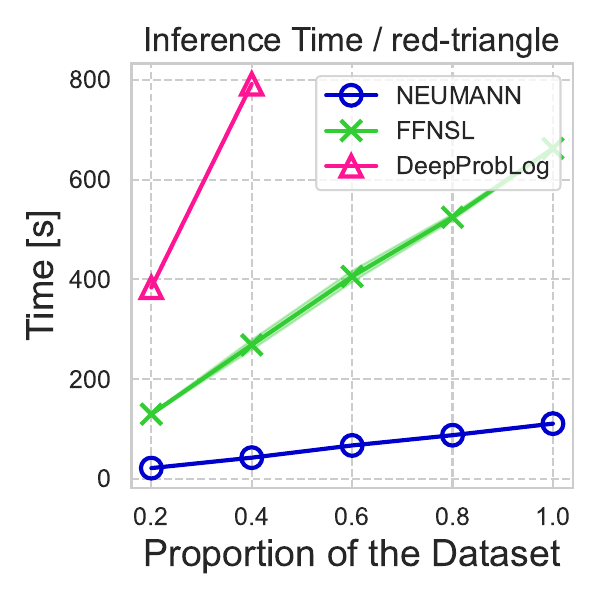} \\
\includegraphics[width=0.31\textwidth]{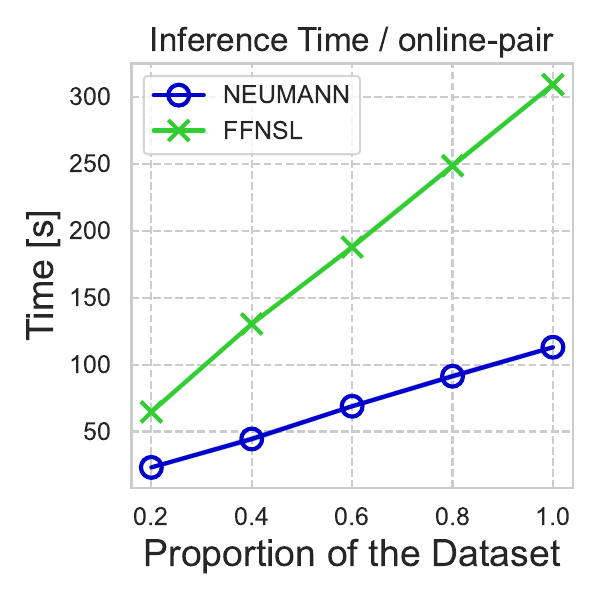} &
\includegraphics[width=0.31\textwidth]{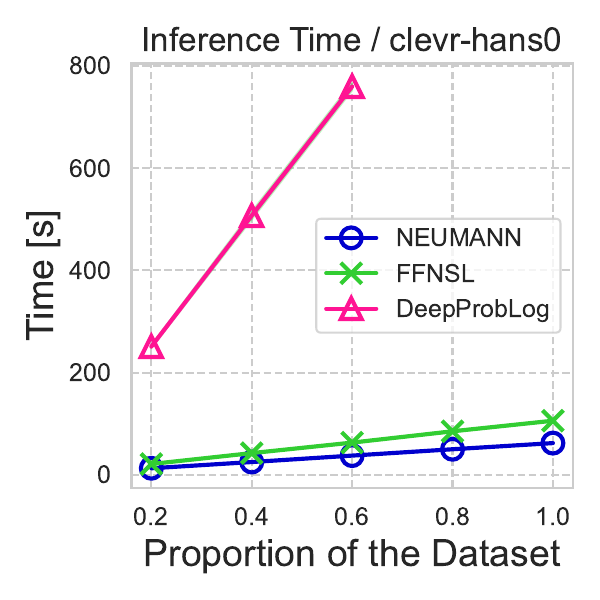} &
\includegraphics[width=0.31\textwidth]{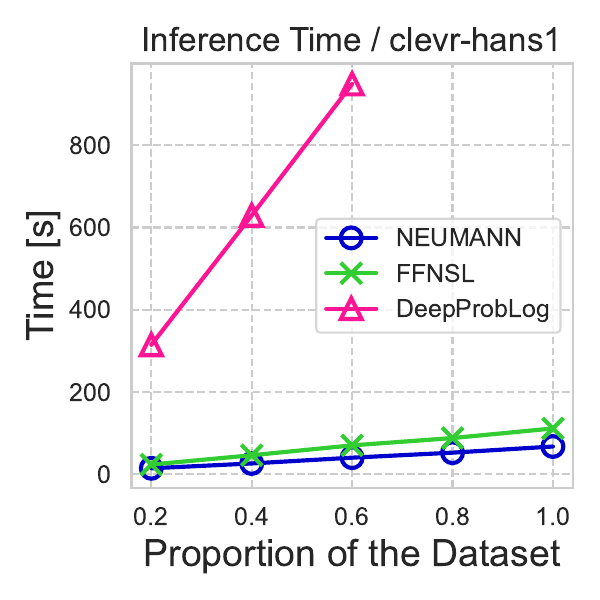} 
\end{tabular}
\caption{\textbf{NEUMANN performs scalable visual reasoning.} We compare the inference time of NEUMANN, FFNSL~\cite{CunningtonLLR23ffnsl} (using CLINGO~\cite{gebser_kaminski_kaufmann_schaub_2019clingo}), and DeepProbLog~\cite{Manhaeve19} (using ProbLog~\cite{de2015probabilistic}) on Kandinsky and CLEVR-Hans using the training split (20k images for each Kandinsky pattern, and 9k images for each CLEVR-Hans class). NEUMANN and DeepProbLog compute probabilistic output, and FFNSL computes discrete output. As we mentioned in main text, the time-out was set to 1000 seconds. In online-pair, DeepProbLog timed out even with 20\% of training data.
Computed for 5 different random seeds. (Best viewed in color)}
\label{fig:inference_time_experiment}
\end{figure}

Fig.~\ref{fig:inference_time_experiment} shows the inference-time comparison of NEUMANN, FFNSL, and DeepProbLog on Kandinsky patterns and CLEVR-Hans.
In each dataset, NEUMANN achieved the fastest inference among the baselines.
Especially in the patterns that require complex classification rules and many ground atoms, \eg red-triangle and online-pair, NEUMANN significantly outperformed DeepProbLog and FFNSL. 
On online-pair, DeepProbLog timed out even with $20\%$ of training data.
This shows that NEUMANN can perform scalable visual reasoning for a large amount of data. 
We note that NEUMANN and DeepProbLog achieve differentiable reasoning, \ie we can obtain gradients out of the probabilistic reasoning result, but FFNSL does not provide this function because of its pure-symbolic reasoner.
 
\begin{figure}[t]
\centering
\begin{tabular}{cc}
\includegraphics[width=0.4\textwidth]{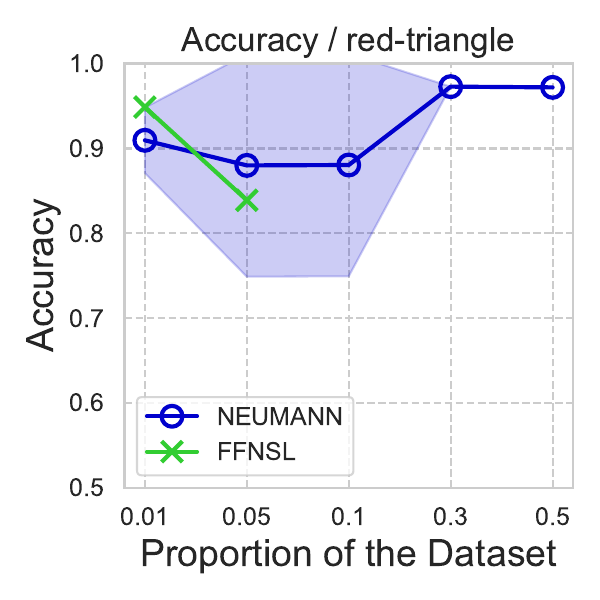} &
\includegraphics[width=0.4\textwidth]{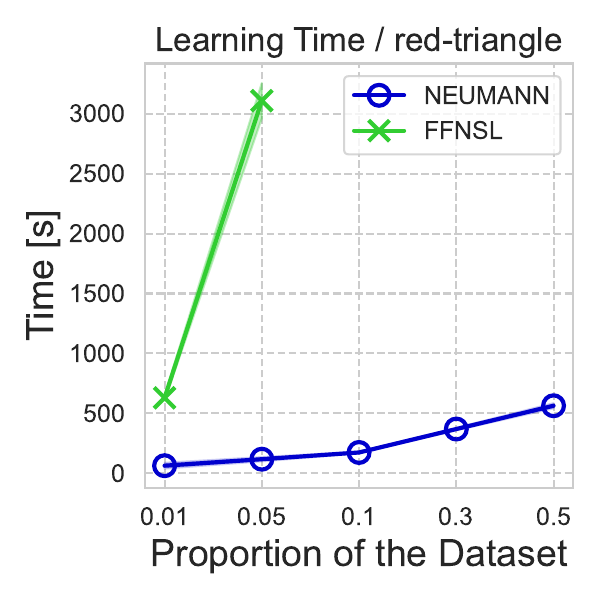} \\
\includegraphics[width=0.4\textwidth]{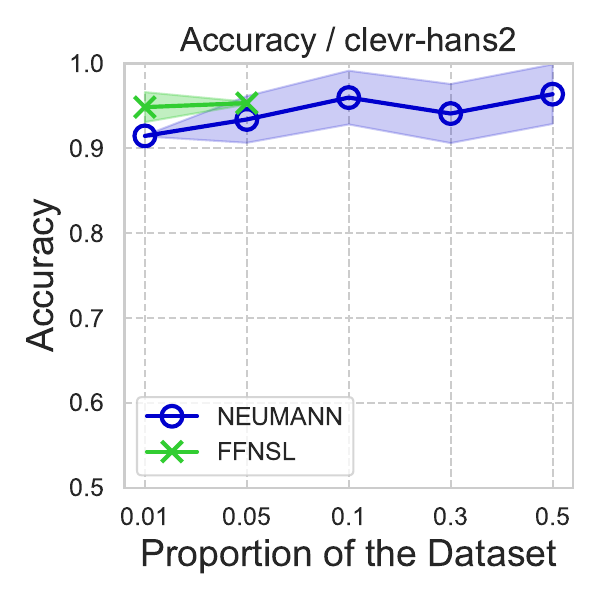} &
\includegraphics[width=0.4\textwidth]{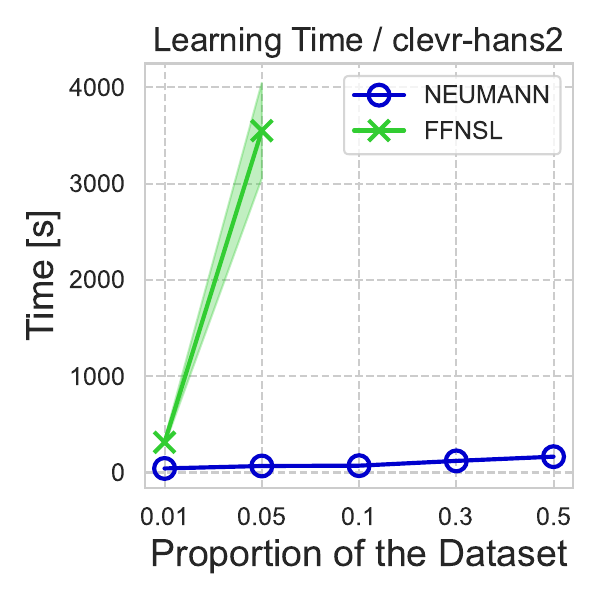} \\
\end{tabular}
\caption{\textbf{NEUMANN performs scalable learning on complex visual scenes.} We compare the performance and learning time of NEUMANN and FFNSL~\cite{CunningtonLLR23ffnsl} (using ILASP~\cite{ILASP}, which serves as a symbolic-learning benchmark), on Kandinsky patterns and CLEVR-Hans by changing the proportion of training/validation data. The accuracy is measured on the full test data. 
As mentioned in main text, the time-out was set to 5000 seconds and FFNSL timed out with more than 10\% of training data. Computed for 3 different random seeds. (Best viewed in color) }
\label{fig:learning_acc_time}
\end{figure}

Moreover, to show that NEUMANN performs scalable visual learning, we compare the performance and running time of learning of NEUMANN and FFNSL using Kandinsky and CLEVR-Hans.
In this setting, FFNSL serves as a \emph{symbolic} learning benchmark because it uses symbolic ILP systems, \eg ILASP~\cite{ILASP}, to learn programs.

We changed the proportion of the training and validation data: 0.01 (1 \%), 0.05 (5 
 \%),  0.1 (10 \%), and 0.5 (50 \%). 
 We measured the test accuracy using the full test data (100 \%) consistently for each setting.
 We measured the running time of learning, which includes the whole process, \ie visual perception, obtaining logic representations, and search logic programs.
We used a batch size of 200 and a learning rate of $1e-2$, and trained 20 epochs.
 We set $(N_\mathit{trial}, N_\mathit{sample})$ as $(4, 10)$ for red-triangle in Kandinsky and $(6, 10)$ for the third pattern of CLEVR-Hans.
 To achieve FFNSL on Kandinsky and CLEVR-Hans, we convert each visual scene to a weighted example for ILASP as described in~\cite{CunningtonLLR23ffnsl}.
 We set to time out as $5000$ seconds. %

Fig.~\ref{fig:learning_acc_time} shows the accuracy of the test split (left) and the learning time (right).
In both datasets, FFNSL achieved high accuracy with a very small number of training data (1\% of training visual scenes), showing the advantage of the symbolic learning method to generalize from small data. However, as the number of training data increases, the learning time increases drastically. To this end, FFNSL handled less than 10\% of the training visual scenes in Kandinsky and CLEVR-Hans.
In contrast, NEUMANN performed learning much faster than FFNSL using a large number of training visual scenes. This shows that NEUMANN is scalable for large datasets, \ie it can learn from a large number of visual scenes. 
Moreover, in Kandinsky, although NEUMANN produced test accuracy with a large variance with a small number of training data, it achieved higher accuracy and gained stability with less variance by using more data.
Overall, NEUMANN outperformed FFNSL in terms of learning time, allowing the system to use more training data and producing competitive test accuracy.

\subsubsection{Explanations by Gradients}
\label{sec:explanation}
We show that NEUMANN can produce gradient-based explanations efficiently by working with neural networks seamlessly.
We use \emph{input gradients}~\cite{Baehrens10inputgradient}, which is a widely-used explanation method for any differentiable models.
Input gradients are gradients with respect to input, \ie for a differentiable function $f$ and an input $\mathbf{x}$ and output $y$, it computes $\mathbf{e} = \partial y / \partial \mathbf{x}$, where each element $\mathbf{e}$ represents how the corresponding input (\eg a pixel) is effective to the output\footnote{For simplicity, we do not take the element-wise product $\mathbf{x} \odot \frac{\partial y}{\partial \mathbf{x}}$.}.
NEUMANN can produce input gradients over input atoms, \ie we compute:
\begin{align}
    \mathbf{e}_\mathit{atoms} = \frac{\partial y}{\partial \mathbf{x}_\mathit{atoms}^{(0)}} =  \frac{\partial \mathbf{x}_\mathit{atoms}^{(T)}}{\partial \mathbf{x}_\mathit{atoms}^{(0)}} \cdot  \frac{\partial y}{\partial \mathbf{x}_\mathit{atoms}^{(T)}}.
    \label{eq:atom_gradients}
\end{align}
Since the reasoning function of NEUMANN is end-to-end differentiable, $\mathbf{e}_\mathit{atoms}$ can be computed efficiently using automatic differentiation (AD), \ie just calling the backward function once after the reasoning\footnote{Since $\mathbf{x}^{(0)}_\mathit{atom}$ is not a leaf node in the computational graph, we prepare a dummy variable $\mathbf{z}_0$ with the same shape as $\mathbf{x}_\mathit{atoms}^{(0)}$ and all elements are initialized with $0$. In the forwarding, we compute $\mathbf{x}^{(0)}_\mathit{atoms} = \mathbf{x}^{(0)}_\mathit{atoms} + \mathbf{z}_0$, and extract gradients stored in $\mathbf{z}_0$ after calling the backward function.}.

Let $\mathbf{M}_1, \ldots, \mathbf{M}_n$ be (attention) masks over $n$ objects for an input scene produced by an object-centric  perception network.
We compose visual explanations as follows:
\begin{align}
    E(\mathbf{x}) = \sum_i\phi_i \cdot \mathbf{M}_i 
    \label{eq:heatmap}
\end{align}
where $\phi_i$ is a weight for the $i$-th mask, which is computed as the maximum probability of input ground atoms regarding the $i$-th object.
Let $\mathcal{J} = \{ j_1, \ldots, j_m \}$ be indices of the ground atoms regarding the $i$-th object (\emph{e.g.} $\mathtt{color(obj_i,red)}$) in ordered set of ground atoms $\mathcal{G}$.
We compute the weight $\phi_i$ for the mask $\mathbf{M}_i$:
\begin{align}
    \phi_i = \max (\mathbf{e}_\mathit{atoms}[j_1], \ldots, \mathbf{e}_\mathit{atoms}[j_m]),
\end{align}
where $\mathbf{e}_\mathit{atoms}$ is computed by Eq.~\ref{eq:atom_gradients}.
For example, if $\mathtt{obj1}$ is a key factor for the classification, the corresponding attention mask $\mathbf{M}_1$ gets a large weight, \ie highlights $\mathtt{obj1}$.%
To this end, Eq.~\ref{eq:heatmap} computes a heatmap that highlights only objects which are effective in the reasoning result.

\begin{figure}[t]
\centering
\includegraphics[width=.99\linewidth]{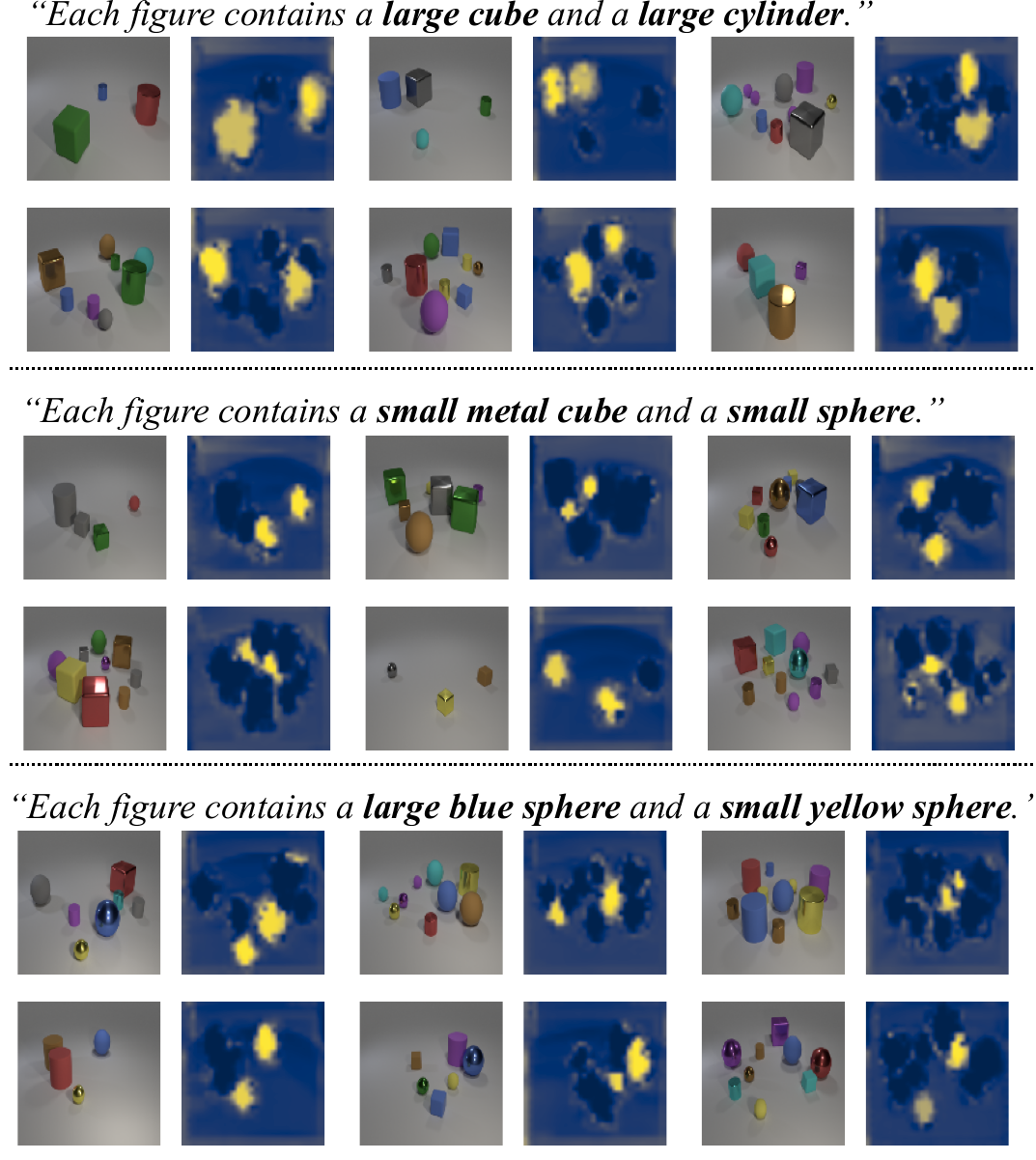}
\caption{\textbf{NEUMANN can explain its reasoning using gradients.} Explanations produced by NEUMANN by computing input gradients~\cite{Baehrens10inputgradient} for input probabilistic atoms $\mathbf{x}^{(0)}_\mathit{atoms}$ and visualizing together with relevant attention maps.  (Best viewed in color)}
\label{fig:clevr-hans_explanation}
\end{figure}

Fig.~\ref{fig:clevr-hans_explanation} shows explanations produced by NEUMANN for CLEVR-Hans using clauses listed in Fig.~\ref{fig:clauses_kandinsky_clevr}.
For each pair of images, the left one shows the original input and the right one shows the heatmap.
NEUMANN successfully produced explanations highlighting objects which are the factors for the classification. 
For example, the first class is about a \emph{large cube} and a \emph{large cylinder}, and NEUMANN highlights both but not others for each input. The explanation can be completed very efficiently by using automatic differentiation (AD).

The same explanation could be obtained by using gradients in DeepProbLog. However, as shown in Section~\ref{sec:inference_time_comparison}, it has a scalability issue for a large amount of complex visual scenes. In contrast, NEUMANN produces gradient-based explanations but still can handle a large amount of complex visual scenes in a scalable manner.
Moreover, FFNSL relies on the discrete inference engine, CLINGO, and thus it is difficult to produce gradient-based explanations using automatic differentiation, \ie it requires additional hard coding for explanations.
Overall, NEUMANN is the only framework that achieves scalable differentiable forward reasoning and learning, producing explanations on complex visual scenes working with neural networks efficiently.

\subsection{Discussions}
We now discuss NEUMANN's advantages, computation, impact, and limitations.

\noindent
\textbf{\emph{What are the advantages compared to pure-symbolic learners?}}

The most promising feature of NEUMANN compared to pure symbolic systems is its capability to handle a large amount of visual input in a scalable manner. 
As shown in Sec.~\ref{sec:comparison_to_symbolic}, NEUMANN can perform visual reasoning and learning, outperforming state-of-the-art neuro-symbolic benchmarks regarding running time and performance.
This feature is crucial for tightly integrating learning and reasoning with neural networks, \eg \emph{algorithmic supervision}~\cite{Petersen21algorichmic_supervision} where neural networks are trained efficiently using gradients via symbolic algorithms.
For such a setting, reasoners should be able to conduct scalable reasoning for a large amount of data to train neural networks. Otherwise, they would be the bottleneck, limiting the applicability of the neuro-symbolic systems.

Moreover, as shown in Sec.~\ref{sec:explanation}, NEUMANN can use gradient-based XAI methods to produce visual explanations working with perception networks efficiently, and it is difficult to produce the same result with a pure symbolic system without additional hard coding.
This feature of NEUMANN leads to essential applications, \eg the \emph{right for the right reasons}~\cite{ijcai2017rightfortherightreasons} approach, which trains neural networks to produce correct explanations using the gradient-based explanations.

Overall, NEUMANN does not contradict symbolic approaches but provides a basis for better neuro-symbolic systems. It has been reported that a pure symbolic system can handle noisy examples~\cite{law2018noisy_asp}, refuting the motivation of $\partial$ILP~\cite{Evans18}. However, NEUMANN provides many other benefits of scalable reasoning and learning paradigm that is compatible with gradient-based methods that include a significant part of the success of DNNs.

\noindent
\textbf{\emph{What makes NEUMANN's reasoning and learning scalable?}}

The scalable performance of NEUMANN can be explained by two reasons. (1) NEUMANN \emph{grounds programs once}, then use the resulting computational graph repeatedly, as other differentiable forward reasoners do~\cite{Evans18,Shindo21,shindo23alphailp}. 
It means that NEUMANN does not compute logic operations (\eg unification) for each specific query.
Instead, NEUMANN performs forwarding on the computational graph and then obtains the results.
In contrast, (differentiable) backward reasoning, employed in DeepProbLog~\cite{Manhaeve19}, needs to construct a new computational graph for a new query, making the reasoning expensive.
(2) More importantly, \emph{NEUMANN is a graph neural network and performs reasoning on GPUs}\footnote{NEUMANN is implemented using PyG, an established GNN library \url{https://pyg.org/}}.
The ground-once scheme enables the reasoner to build and fix the computational graph as users do with neural networks, \ie defining the network architecture and a set of weights, then the computational graph is constructed and fixed so forwarding can be conducted on GPUs.
When dealing with a batch of examples (\eg 200 examples), NEUMANN can process them in parallel very efficiently.
This feature is not trivial for logic reasoners. Typically, they process a batch of examples sequentially.
For instance, DeepProbLog uses Sentential Decision Diagrams~\cite{Darwiche11SDD} for its reasoning, and it requires building different SDDs on CPUs for each query, and FFNSL uses the CPU-based reasoner (CLINGO~\cite{gebser_kaminski_kaufmann_schaub_2019clingo}). Thus it requires non-trivial efforts for these reasoners to compute reasoning by using GPUs in a scalable manner.

\noindent
\textbf{\emph{Why is it crucial to improve the memory consumption of differentiable forward reasoners? }}

Differentiable forward reasoning, introduced in the $\partial$ILP framework~\cite{Evans18}, encodes logic programs to \emph{tensors}. Differentiable forward reasoning is inherently memory intensive and thus limits the expressivity of logic programs, \eg no functors are allowed, each rule consists of at most 2 body atoms, and each predicate takes at most 2 arguments.
$\partial$ILP has been extended to handle structured programs with functors~\cite{Shindo21}, by incorporating search techniques in ILP~\cite{Muggleton95,Nienhuys97,Cropper20}, leading to $\alpha$ILP~\cite{shindo23alphailp}, which learns logic rules from complex visual scenes. However, these methods inherit the memory-intensive tensors and thus cannot handle complex programs for abstract visual reasoning.
Differentiable forward reasoning gains several advantages compared to pure symbolic reasoners, as discussed in the previous paragraph. Thus it is crucial to improve memory consumption so that a wide range of programs can be handled in the framework to expand the applicability of neuro-symbolic approaches.

\noindent
\textbf{\emph{Why is graph encoding more efficient than dense tensor encoding?}}

Graphs can represent the relations between different atoms more efficiently than dense tensors.
For example, suppose we want to encode the following information, \emph{"Atom $X$ is not deduced by atom $Y$"}. A reasoning graph can represent this information simply by not having edges between $X$ and $Y$. In contrast, the dense tensors in differentiable ILP frameworks~\cite{Evans18,Shindo21,shindo23alphailp} need to hold \emph{false symbol} $\bot$ for each combination of $X$ and $Y$, \ie the system needs to keep all of the combinations of ground atoms. %
This results in NEUMANN being memory efficient as shown in Proposition~\ref{prop:memory}.

\noindent
\textbf{\emph{Can sparse tensors result in reasoners comparable to message-passing reasoners?}}

Sparse tensors cannot account for differentiable forward reasoning because they do not support some essential tensor operations %
as they compress the tensors by breaking the row-and-column structure. As shown in $\partial$ILP~\cite{Evans18}, differentiable forward reasoning uses the \emph{slice} and \emph{gather} operations on tensors repeatedly, and those are not supported by sparse tensors.
Thus simply using sparse tensors does not lead to memory-efficient differentiable forward reasoners.

\noindent
\textbf{\emph{Why can existing VQA models not solve the Behind-the-Scenes task?}}

VQA models accept input as a tuple of an image and a question in natural-language sentences~\cite{Antol15vqa}. 
Symbolic programs, typically described as a Domain Specific Language (DSL) to compute answers, are generated by parsing the input question using neural networks.
Given visual scenes, it is unclear how to perform program induction on their DSL since there is no uniform structure-learning algorithm for each DSL.
Thus simply using existing VQA models for the proposed task cannot be a solution.

\noindent
\textbf{Limitations.}
Although NEUMANN is a more general framework compared to classic symbolic and neuro-symbolic frameworks, it does suffer from some limitations: (1) The language to be handled is limited to definite clauses, which are rules in FOL with a single head atom. Symbolic systems can handle more complex structures, \eg choice rules in Answer Set Programming (ASP) systems~\cite{BrewkaET11ASP}.
(2) The learning algorithm is not jointly training perception networks and logic programs.
(3) The message-passing algorithm is not connected to well-known probabilistic semantics, \eg distribution semantics~\cite{Sato95distributionsemantics}.
\section{Related Work}
NEUMANN builds upon different sub-fields of AI.
We revisit relevant studies.

\textbf{Symbolic AI.}
Symbolic representations, \eg First-Order Logic (FOL), provide essential functions of knowledge representation and reasoning capabilities to AI systems, which are difficult to be provided by purely neural-based models~\cite{baral2003knowledge,brachman2004knowledge,nickel2015review}. 
A  pioneering study of inductive inference was done in the early 70s~\cite{Plotkin71}.
Many systems have been developed for \emph{inductive inference}~\cite{Angluin83}, \eg Model Inference System (MIS)~\cite{Shapiro83}  has been implemented as an efficient search algorithm for logic programs.
Inductive Logic Programming (ILP)~\cite{Muggleton95,Nienhuys97,Cropper20} has emerged at the intersection of machine learning and logic programming. 
ILP systems using Answer Set Programming (ASP) can learn logic programs beyond definite clauses~\cite{ILASP,fastlas}, \eg choice rules.
ILP has advantages compared to data-driven DNNs, \eg it can learn explicit programs and learn from small data. Thus, combining ILP with DNNs is a promising approach to overcoming the limitations of the current data-driven machine-learning paradigm.
NEUMANN embraces the symbolic learning approaches in the neuro-symbolic setting, where logical reasoning and neural learning are tightly integrated.

\textbf{Probabilistic Logic and Neuro-Symbolic AI.}
Combining probabilities with symbolic logic has been addressed to establish reasoning systems that can handle uncertainty, \eg distribution semantics ~\cite{Sato95distributionsemantics} and Bayesian logic programs~\cite{kersting2008bayesianlogic}.
Probabilistic Inductive Logic Programming~\cite{DeRaedt08probabilisticilp} combines ILP with probabilistic semantics establishing a new learning paradigm.
Structure learning algorithms for probabilistic logic programs have been developed, \eg SLIPCOVER~\cite{bellodi_riguzzi_2015slipcover}.
These approaches focus on learning with probabilistic semantics, but NEUMANN engages differentiable reasoning and learning, where parameters get gradients optimized via gradient descent.
Lifted inference~\cite{LiftedBook21} addresses efficient reasoning, \eg reducing computational graphs by using symmetry, and these techniques could be incorporated into NEUMANN since it employs graphs as its representation.
Markov Logic Networks (MLNs)~\cite{Richardson07} takes a similar approach to ground the logic programs to produce a graph structure. MLNs perform inference based on Bayesian networks, but NEUMANN computes differentiable forward reasoning by message-passing as graph neural networks.

Integration of symbolic computations and neural networks, called \emph{neuro-symbolic AI}~\cite{Garcez20thirdwave,kautzthird}, has attracted a lot of attention in recent years. Many frameworks have been developed for parameter estimation of DNNs using symbolic programs, \eg DeepProbLog~\cite{Manhaeve19,Manhaeve21}, NeurASP~\cite{Yang20}, SLASH~\cite{SkryaginKR22slash}, NS-CL~\cite{Mao19}, differentiable theorem provers~\cite{rocktaschel17}, and Embed2Sym~\cite{Aspis22kr}. In a similar vein, differentiable structure learners have been developed~\cite{Evans18,minervini20icml,Shindo21,shindo23alphailp}, and NEUMANN extends their capacity by having memory-efficient reasoning and computationally-efficient learning.
TensorLog~\cite{cohen20} performs message-passing for backward reasoning, but NEUMANN realizes forward reasoning.
GNNs have been used for reasoning~\cite{Rawson20graph,Paliwal20} by composing logical expressions as graphs, where neural representations are trained given symbolic knowledge. In contrast, NEUMANN performs structure learning using reasoning graphs and fuzzy logic operations.  %
Logical Neural Networks (LNNs)~\cite{Riegel20} is a class of neural networks where each node has its logical semantics. 
LNNs parameterize soft-logical operations, but NEUMANN parameterizes clauses with their weights.
Lifted Relational Neural Networks~\cite{Sourek18} uses rules as a template to produce deep neural networks. NEUMANN uses rules as a template to produce differentiable message-passing forward reasoner.
Integration DNNs with abductive learning~\cite{Dai19} has been addressed, where the agent learns to complete a symbolic knowledge base.
This approach does not address program induction from raw inputs. In contrast, NEUMANN performs structure learning from complex visual scenes.
MetaABD~\cite{DaiM21MetaABD} has been proposed to perform program induction based on abductive learning by integrating a learning system Metagol~\cite{metagol}.
Metagol handles definite clauses without functors, but NEUMANN can learn programs with functors. %

\textbf{Visual Reasoning Datasets.}
The deep-learning community has developed many image datasets for evaluating different image-recognition models, \eg MNIST~\cite{deng2012mnist} and ImageNet~\cite{deng09imagenet}. However, these datasets are dedicated to simple label classification, and thus difficult to assess the reasoning abilities of machine-learning models.
To overcome this limitation, visual datasets with reasoning requirements have been developed.
Visual Question Answering (VQA)~\cite{Antol15vqa,wu2017visualquestionanswering,krishna16visualgenome} is a well-established scheme to learn to answer questions given as natural language sentences together with input images.  %
VQA has an assumption that the programs to compute answers are given as input questions. However, in Behind-the-Scenes, the agents need to learn abstract programs to compute the answers. Moreover, VQA models do not address learning from small data and transferring the obtained knowledge to new tasks, which are parts of the Behind-the-Scenes requirement.
Neuro-symbolic models achieve multi-modal learning on VQA~\cite{yi2018neural,Tan19multimodal}, but NEUMANN addresses rather structure-learning problems with differentiable logic programming.
VQAR~\cite{Huang21scallop} is a variant of VQA with relational reasoning, CLEVRER~\cite{Yi2020CLEVRER} is an extension of CLEVR with sequential input, and MNIST-Addition~\cite{Manhaeve19} is about learning DNNs to add hand-written digits.
These datasets involve essential aspects of reasoning, \eg, relations with multiple entities, temporal reasoning, and arithmetic computation. However, as shown in Tab.~\ref{tab:compare_tasks}, Behind-the-Scenes achieve the \emph{four} essential model-building aspects: (i) learning from small data, (ii) learning from complex visual scenes, (iii) learning explanatory programs, and (iv) reasoning beyond observations. 
Previously proposed datasets cover some of these aspects but not all of them. The proposed Behind-the-Scenes task is the first dataset to assess the four model-building abilities of machine-learning models.
Abstract Visual Reasoning (AVR) has been addressed to test the ability to apply previously gained knowledge and programs in a completely new setting, posing challenges to DNNs~\cite{Sheng21AVR,milkolaj23AVR,Camposampiero23AVR}.
The methods have been evaluated on simple tasks of abstract puzzles, \eg Raven's progressive matrices~\cite{raven1998raven}.
The proposed task, Behind-the-Scenes, requires structured program induction and reasoning beyond observation in complex 3D visual scenes, which have not been addressed in previous AVR studies.

A motivation of the proposed behind-the-scene task is \emph{problem solving},
which is an essential aspect of human intelligence of solving problems beyond perception using reasoning~\cite{NewellSimon1961humanthinking,Newell1972humanproblemsolving}.
Humans can learn much from a small number of experiences developing capacities to represent physical objects and reason about their motion~\cite{Spelke92origins,Battaglia13physics}.
Inspired by these studies, the development of adaptive learning skills of humans has been addressed as \emph{model building problems}~\cite{Lake17buildingmachines}, and data-driven DNNs are insufficient to achieve these aspects.
NEUMANN tackles this challenge by performing memory-efficient differentiable forward reasoning using DNNs.

\textbf{Graphs and Circuits for Reasoning.}
Many approaches have been developed to encode the reasoning structures to graphs and circuits.
Binary Decision Diagrams (BDDs)~\cite{Bryant86BDD} encode propositional logic expressions compactly as a directed graph, leading to variant structures, \eg Sentential Decision Diagrams (SDDs)~\cite{Darwiche11SDD,KisaKR14PSDD}, Zero-suppressed Decision Diagrams (ZDDs)~\cite{Minato93ZDD}, and Zero-supressed Sentential Decision Diagrams (ZSDDs)~\cite{Nishino_Yasuda_Minato_Nagata_2016ZSDD}. These architectures are developed for propositional logic or combinatorial optimization, but the reasoning graph of NEUMANN represents first-order logic and addresses differentiable reasoning. Their efficient compression algorithms and operations between graphs (\eg taking conjunction between two graphs) for these structures could be applied to reasoning graphs in NEUMANN.
Sum-Product Networks (SPNs)~\cite{PoonD11SPN} encode tractable probability distributions in graphs, which repeatedly consist of sum and product layers. NEUMANN shares a similar structure since its reasoning graph consists of atom nodes to compute disjunctions and conjunction nodes, and performs bi-directional message-passing. SPNs solve exact probabilistic inference, but NEUMANN addresses differentiable reasoning on first-order logic.

\section{Conclusion}
We presented NEUMANN, a memory-efficient differentiable forward reasoner that passes messages on reasoning graphs. 
NEUMANN compiles logic programs in first-order logic to a graph that encloses the process of forward reasoning and performs message-passing in a neural fashion. 
Moreover, we proposed a computationally-efficient learning algorithm combining gradient-based scoring and differentiable sampling of clauses.
In our experiments, we have shown: 
(1) The message-passing reasoning algorithm simulates the differentiable forward reasoning dealing with uncertainty.
(2) NEUMANN can solve visual ILP problems combined with DNNs, outperforming neural baselines and consuming less memory than the other differentiable ILP benchmarks.
(3) NEUMANN solves the Behind-the-Scenes task outperforming conventional differentiable reasoners, providing model-building abilities beyond simple perception capabilities, \ie learning from small data, understanding visual scenes deeply, learning explanatory programs, and reasoning about non-observational scenes.
(4) NEUMANN performs scalable visual reasoning and learning, outperforming state-of-the-art symbolic and neuro-symbolic methods regarding running time and performance. Moreover, NEUMANN can incorporate XAI methods efficiently, \ie NEUMANN produces gradient-based visual explanations using DNNs.

NEUMANN provides several interesting avenues for future work. 
NEUMANN is an instance of GNNs, providing the capability of representation learning to make neuro-symbolic reasoning more robust and multi-modal.
Moreover, NEUMANN enables differentiable reasoning on complex logic programs with functors and thus can be used for vital applications, such as planning, meta-interpreters, and knowledge-enhanced foundation models. 
NEUMANN is also promising for the \emph{right for the right reasons} approach~\cite{ijcai2017rightfortherightreasons}, where neural networks are trained to produce correct explanations and thus a vital factor to achieve explainable machine learning systems.
Generally, it bridges the current data-driven machine learning paradigm to perform problem-solving beyond perception with knowledge representation and reasoning.

\bibliographystyle{abbrv}
\bibliography{sn-article}

\newpage
\appendix
\section{Logic Programs for Behind-the-Scenes}
We show logic programs used for solving the Behind-the-Scenes task but not shown in the main text.
Fig.~\ref{fig:clauses_utility} shows a set of clauses to define utility predicates to solve the task, \eg extracting a color of an object according to its position.
Fig.~\ref{fig:append_reverse} shows additional logic programs used for query answering. 
\begin{figure}[h]
\centering
\begin{lstlisting}[language=Prolog,  style=Prolog-pygsty]
% extract the target color according to the position
% e.g. get the color of the 1st (left-most) object
get_color([Color1,Color2,Color3],Position,X):-
                first_obj([Color1,Color2,Color3],X),
                same_position(Position,1st).
get_color([Color1,Color2,Color3]),Position,X):-     
                second_obj([Color1,Color2,Color3]),X),
                same_position(Position,2nd).
get_color([Color1,Color2,Color3],Position,X):-
                third_obj([Color1,Color2,Color3],X),
                same_position(Position,3rd).
                
% mapping the list of colors to a color, according to the position
% used by the get_color predicate
first_obj([Color1,Color2,Color3],Color1):-.
second_obj([Color1,Color2,Color3],Color2):-.
third_obj([Color1,Color2,Color3],Color3):-.
\end{lstlisting}
    \caption{Additional clauses to define utility predicates to answer queries.}       
    \label{fig:clauses_utility}
\end{figure}

\begin{figure}[h]
        \centering
           \begin{lstlisting}[language=Prolog,  style=Prolog-pygsty]
% append(X,Y,Z) means Z is the result of appending X to the head of Y.
append([],X,X):-.
append([X|Y],Z,[X|V]):-append(Y,Z,V).

% reverse(X,Y) means that Y is the result of reversing X.
reverse([H|T],A,R):-reverse(T,[H|A],R).
reverse([],A,A):-.

% permutation(X,Y) means Y is a permutation of X.
permutation([X,Y],Z):-permutation(Y,V),delete(X,Z,V).
\end{lstlisting}
    \caption{Additional clauses to compute abstract list operations.}
    \label{fig:append_reverse}
\end{figure}

\section{Experimental Details on Reasoning Behind-the-Scenes}
\label{sec:detail_behind-the-scenes}
We describe the experimental details of the behind-the-scenes task.

\textbf{Task 1 (T1).}
We used the mode declarations in Tab.~\ref{tab:mode_bs} for both NEUMANN and $\partial$ILP-ST. The definition of mode declarations is available in Sec.~\ref{sec:definition_mode_declarations}.
We performed $50$ epochs of weight optimization with a batch size of $64$.
We used the RMSProp~\cite{Ruder16gradient} optimizer with a learning rate of $1e-2$.
To prune too general clauses in the clause generation step, we gradually increased the ratio of negative examples from 20\% to 100\% by 20\% in the first 5 trials of the clause generation. 
For $\mathtt{sort}$, we performed curriculum learning, \eg learning a simple predicate first ($\mathtt{is\_sorted}$) and then finalizing the complete learning ($\mathtt{sort}$).

We limit the number of nested functors at most $3$ and discard lists with duplicated elements.
We used the same perception model as in CLEVR-Hans, which is described in Section \ref{sec:perception}.
Visual examples for each operation are shown in Sec.~\ref{sec:more_examples}.

\textbf{Task 2 (T2).}
Queries about different operations are given randomly, so the model needs to handle different types of queries in the prediction.
To generate visual scenes with queries and their answers, we adopted the generation code of CLEVR~\cite{Johnson17}.
When solving, NEUMANN reads out a JSON file which contains instances, and each instance contains a path to an image file and pairs of a query and a corresponding answer, \eg $(\mathtt{query2(q\_delete,cyan,}$ $\mathtt{2nd)}, $ $\mathtt{ans(red)})$.
NEUMANN assigns probabilities over query atoms according to the input, \ie it gives $1.0$ for a query atom given input and $0.0$ for other query atoms.
The answer is used only for computing accuracies and never in the prediction pipeline. 

We used a batch size of $64$ for NEUMANN and $1$ for $\alpha$ILP, \ie $\alpha$ILP produced out-of-memory even with the smallest batch size, as shown in Tab.~\ref{tab:acc_behind-the-scenes}.
We used the same perception model as in CLEVR-Hans, which is described in Section \ref{sec:perception}.
We limit the number of nested functors at most $3$ and discard lists with duplicated elements.
We used additional abstract operations ($\mathtt{append}$ and $\mathtt{reverse}$) shown in Fig.~\ref{fig:append_reverse}.
Examples of input scenes paired with queries and their answers are shown in Sec.~\ref{sec:more_examples}.

\begin{table}[t]
\centering
    \caption{Mode declarations for Behind-the-Scenes.}
    \begin{tabular}{l}
    \hline
    ${\tt modeb(1, member(+object, +objcets))}$\\
    ${\tt modeb(1, delete(+object, +objcets, +objects))}$\\
    ${\tt modeb(1, is\_sorted(+objects))}$\\
    ${\tt modeb(1, smaller(+object, +object))}$\\
    ${\tt modeb(1, permutation(+objects, +objcets))}$\\
    \hline
    \end{tabular}
    \label{tab:mode_bs}
\end{table}

\section{Experimental Details on Kandinsky Patterns and CLEVR-Hans}
\label{sec:detail_kandinsky_clevrhans}
\noindent
In this section, we describe the experimental setting of Kandinsky Patterns and CLEVR-Hans.
\subsection{Kandinsky Patterns}

\textbf{CNN.}
We trained ResNet18 for 300 epochs with a batch size of $512$.
We used the Adam optimizer~\cite{adam,Ruder16gradient} with a learning rate of $1e-5$.

\noindent
\textbf{YOLO+MLP.}
We used MLP with \emph{two} hidden layers. Each hidden layer applies a linear transformation and a non-linearity.
The output of the pre-trained YOLO model is reshaped and fed into MLP to predict the class label.
We jointly trained the whole YOLO+MLP network for $1000$ epochs with a batch size of $512$. 
We used the Adam optimizer~\cite{adam,Ruder16gradient} with a learning rate of $1e-5$.

\noindent
\textbf{$\alpha$ILP/NEUMANN.}
We trained the $\alpha$ILP and NEUMANN model for 100 epochs with a batch size of $64$.
We used the RMSProp~\cite{Ruder16gradient} optimizer with a learning rate of $1e-2$.
For $\alpha$ILP, we used $500$ positive examples in the validation split to generate clauses by beam search.

Mode declarations we used are shown in Tab. \ref{tab:mode_kp}.
Tab. \ref{fig:const_kandinsky} shows the data types and constants, and Tab. \ref{tab:pred_kandinsky} shows the predicates used in our experiments.
\#obj represents the number of objects to be focused on the classification, which can be identified by trying from the smallest number and evaluating by validation split and increasing if the performance is not enough. 
We set the initial clause to be the root node in the beam search as:
${\tt kp(X)} ~\mbox{:-}~ {\tt in(O1,X),\ldots,in(On,X).}$, 
where $n$ is the number of objects to be focused. %
Background knowledge given in for Kandinsky patterns is shown in Tab. \ref{tab:bk_kp}.

\begin{table}[t]
\begin{minipage}{.5\linewidth}
    \centering
    \caption{Mode declarations for Kandinsky patterns and CLEVR-Hans.}
    \begin{tabular}{l}
    \hline
${\tt modeh(1, kp(-image))}$\\
    ${\tt modeb(\#obj, in(-object, +image))}$\\
    ${\tt modeb(1, color(+object, \#color))}$\\
    ${\tt modeb(1, shape(+object, \#shape))}$\\
    ${\tt modeb(2, same\_color\_pair(+object, +object))}$\\
    ${\tt modeb(2, same\_shape\_pair(+object, +object))}$\\
    ${\tt modeb(1, diff\_color\_pair(+object, +object))}$\\
    ${\tt modeb(1, duff\_shape\_pair(+object, +object))}$\\
    ${\tt modeb(1, closeby(+object, +object))}$\\
    ${\tt modeb(1, online(+object, \ldots, +object))}$\\\hline
    ${\tt modeh(1, ch(-image))}$\\
    ${\tt modeb(\#obj, in(-object, +image))}$\\
    ${\tt modeb(1, color(+object, \#color))}$\\
    ${\tt modeb(1, shape(+object, \#shape))}$\\
    ${\tt modeb(1, material(+object, \#material))}$\\
    ${\tt modeb(1, size(+object, \#size))}$\\\hline
    \end{tabular}
    \label{tab:mode_kp}
\end{minipage}
\begin{minipage}{.5\linewidth}
\centering
\caption{Datatype and constants in Kandinsky patterns and CLEVR-Hans.}
\begin{tabular}{ll}
{\bf Datatype} & {\bf Terms} \\ \hline 
${\tt image}$ & ${\tt img}$\\
${\tt object}$ & ${\tt obj1}$, ${\tt obj2}$, $\ldots$, ${\tt obj6}$ \\
${\tt color}$ & ${\tt red}$, ${\tt blue}$, ${\tt yellow}$\\
${\tt shape}$ & ${\tt square}$, ${\tt circle}$, ${\tt triangle}$\\\hline
${\tt image}$ & ${\tt img}$\\
${\tt object}$ & ${\tt obj0}$, ${\tt obj1}$, $\ldots$, ${\tt obj9}$ \\
${\tt color}$ & ${\tt cyan}$, ${\tt blue}$, ${\tt yellow}$, ${\tt purple}$, \\ 
& ${\tt red}$, ${\tt green}$, ${\tt gray}$, ${\tt brown}$\\
${\tt shape}$ & ${\tt sphere}$, ${\tt cube}$, ${\tt cylinder}$\\
${\tt size}$ & ${\tt large}$, ${\tt small}$\\
${\tt material}$ & ${\tt rubber}$, ${\tt metal}$\\
\\
\\
\\
\\
\end{tabular}
\label{fig:const_kandinsky}
\end{minipage}

\end{table}

\begin{table*}[t]
\centering
\caption{Predicates in the Kandinsky patterns and CLEVR-Hans.}
\scalebox{0.9}{
\begin{tabular}{ll}
{\bf Predicate} & {\bf Explanation} \\ \hline 
${\tt kp}/(1,[{\tt image}])$ & The image belongs to the Kandinsky pattern.\\
${\tt same\_shape\_pair}/(2,[{\tt object},{\tt object}])$ & The two objects have the same shape.\\
${\tt same\_color\_pair}/(2,[{\tt object},{\tt object}])$ & The two objects have the same color.\\
${\tt diff\_shape\_pair}/(2,[{\tt object},{\tt object}])$ & The two objects have different shapes.\\
${\tt diff\_color\_pair}/(2,[{\tt object},{\tt object}])$ & The two objects have different colors.\\
${\tt in}/(2,[{\tt object},{\tt image}])$ & The object is in the image.\\
${\tt shape}/(2,[{\tt object},{\tt shape}])$ & The object has the shape of the second argument.\\
${\tt color}/(2,[{\tt object},{\tt color}])$ & The object has the color of the second argument.\\
${\tt closeby}/(2,[{\tt object},{\tt object}])$ & The two objects are located close by each other.\\
${\tt online}/(5,[{\tt object}, \ldots,{\tt object}])$ & The objects are aligned on a line.\\ \hline
${\tt ch}/(1,[{\tt image}])$ & The image belongs to the clevr-hans pattern.\\
${\tt in}/(2,[{\tt object},{\tt image}])$ & The object is in the image.\\
${\tt shape}/(2,[{\tt object},{\tt shape}])$ & The object has the shape of the second argument.\\
${\tt color}/(2,[{\tt object},{\tt color}])$ & The object has the color of the second argument.\\
${\tt material}/(2,[{\tt object},{\tt material}])$ & The object has the material of ${\tt material}$.\\
${\tt size}/(2,[{\tt object},{\tt size}])$ & The object has the size of the second argument.
\end{tabular}
}
\label{tab:pred_kandinsky}
\end{table*}

\begin{table}[t]
    \centering
    \begin{tabular}{l}
        \hline
        ${\tt same\_shape\_pair(X,Y):-shape(X,Z),shape(Y,Z),}$\\
  ${\tt same\_color\_pair(X,Y):-color(X,Z),color(Y,Z),}$\\
 ${\tt diff\_shape\_pair(X,Y):-shape(X,Z),shape(Y,W),diff\_shape(Z,W).}$\\
 ${\tt diff\_color\_pair(X,Y):-color(X,Z),color(Y,W),diff\_color(Z,W),}$\\\hline
 ${\tt diff\_color(red,blue), diff\_color(blue,red),}$\\
 ${\tt  diff\_color(red,yellow), diff\_color(yellow,red),}$\\
${\tt  diff\_color(blue,yellow), diff\_color(yellow,blue).}$\\
${\tt diff\_shape(circle,square), diff\_shape(square,circle),}$\\
${\tt diff\_shape(circle,triangle),diff\_shape(triangle,circle),}$\\
${\tt  diff\_shape(square,triangle), diff\_shape(triangle,square).}$\\\hline
    \end{tabular}
    \caption{Background knowledge for Kandinsky patterns.} 
    \label{tab:bk_kp}
\end{table}

\subsection{CLEVR-Hans}
We trained the $\alpha$ILP and NEUMANN model for 100 epochs with a batch size of $256$.
We used the RMSProp optimizer~\cite{Ruder16gradient} with a learning rate of $1e-2$.
For $\alpha$ILP, we used $500$ positive examples in the validation split to generate clauses by beam search.

Mode declarations we used are shown in Tab. \ref{tab:mode_kp}.
Tab. \ref{fig:const_kandinsky} shows the data types and constants, and Tab. \ref{tab:pred_kandinsky} shows the predicates.
We set the initial clause to be the root node in the beam search as:
${\tt ch(X)} ~\mbox{:-}~ {\tt in(O1,X),in(O2,X)}$.
We did not provide any background knowledge for CLEVR-Hans tasks.
We refer to~\cite{Stammer21} for details about CNN, NeSy, and NeSy-XIL benchmarks.

\subsection{Perception Models}
\label{sec:perception}
We describe the experimental setting of the pre-training of the perception models in our experiments.
\subsubsection{YOLO for Kandinsky Patterns}
\noindent
\textbf{Model.}
We used YOLOv5\footnote{https://github.com/ultralytics/yolov5} model, whose implementation is publicly available. We adopted the YOLOv5s model, which has 7.3M parameters.

\noindent
\textbf{Dataset.}
We generated $15,000$ pattern-free figures for training, $5000$ figures for validation. The class labels and positions are generated randomly.
The original image size is $620 \times 620$, and resized into $128 \times 128$.
The label consists of the class labels and the bounding box for each object.
The class label is generated by the combination of the shape and the color of the object, e.g., {\em red circle} and {\em blue square}.
The number of classes is $9$.
Each image contains at least $2$ objects and, at most $10$ objects.

\noindent
\textbf{Optimization.}
We trained the YOLOv5s model by stochastic gradient descent (SGD) for $400$ epochs using the pre-trained weights\footnote{https://github.com/ultralytics/yolov5/releases}.
We used the loss function that approximates detection performance, presented in~\cite{Redmon16}.
We set the learning rate to $0.01$ and the batch size to $64$.
The SGD optimizer used the momentum, which is set to $0.937$.
We set the weight decay as $0.0005$.
We took $3$ warmup epochs for training.

\subsubsection{Slot Attention for CLEVR}
We used the same model and training setup as the pre-training of the slot-attention module in \cite{Stammer21}.
In the preprocessing, we downscaled the CLEVR images to a dimension of $128 \times 128$ and normalized the images to lie between $-1$ and $1$. For training the slot-attention module, an object is represented as a vector of binary values for the shape, size, color, and material attributes and continuous values between $0$ and $1$ for the $x$, $y$, and $z$ positions. We trained the slot attention model with the set prediction architecture following~\cite{Locatello20}, using the loss function, which is based on the Hungarian algorithm.
We refer to~\cite{Stammer21} for more details.

\section{Mode Declaration}
\label{sec:definition_mode_declarations}
Mode Declaration~\cite{Muggleton95} is one of the common language biases for Inductive Logic Programming.
We used mode declaration, which is defined as follows.
A mode declaration is either a head declaration ${\tt modeh(r, p(mdt_1, \ldots, mdt_n))}$ or a body declaration ${\tt modeb(r, p(mdt_1, \ldots, mdt_n))}$,
where ${\tt r}\in \mathbb{N}$ is an integer, ${\tt p}$ is a predicate, and ${\tt mdt_i}$ is a mode datatype. A mode datatype is a tuple ${\tt (pm, dt)}$, where ${\tt pm}$ is a place-marker and ${\tt dt}$ is a datatype.
A place-marker is either $\#$, which represents constants, or
$+$ (resp. $-$), which represents input (resp. output) variables.
${\tt r}$ represents the number of the usages of the predicate to compose a single clause.

\section{More Examples in Datasets}
\label{sec:more_examples}
\begin{figure}
    \centering
    \includegraphics[width=.9\linewidth]{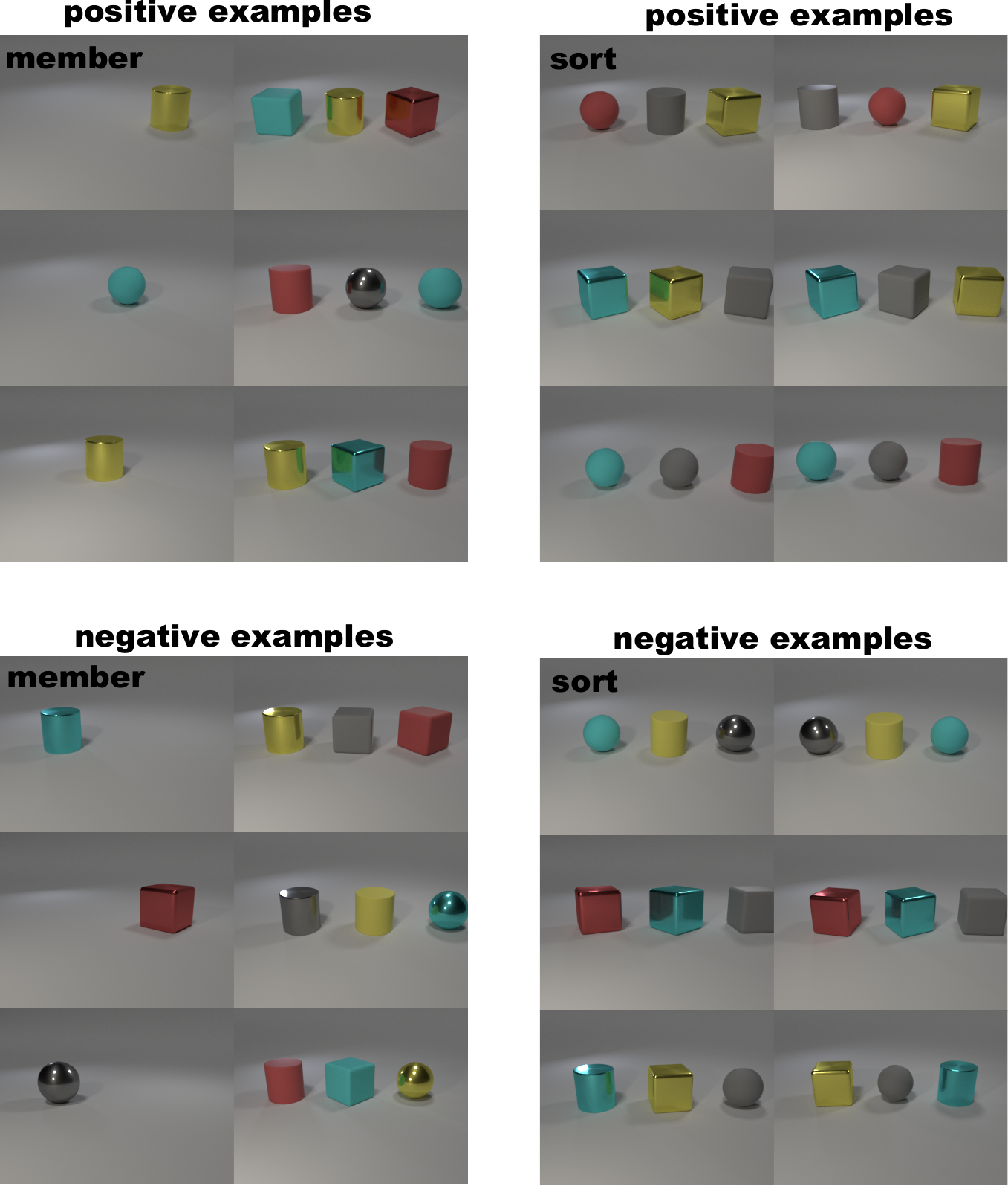}
    \caption{Positive and negative examples for \emph{member} and \emph{sort}  (with an order of colors: \emph{cyan} $<$ \emph{gray} $<$ \emph{red} $<$ \emph{yellow}, alphabetical order) in CLEVR-List.}
    \label{fig:member_sort_examples}
\end{figure}

\begin{figure}
    \centering
    \includegraphics[width=.7\linewidth]{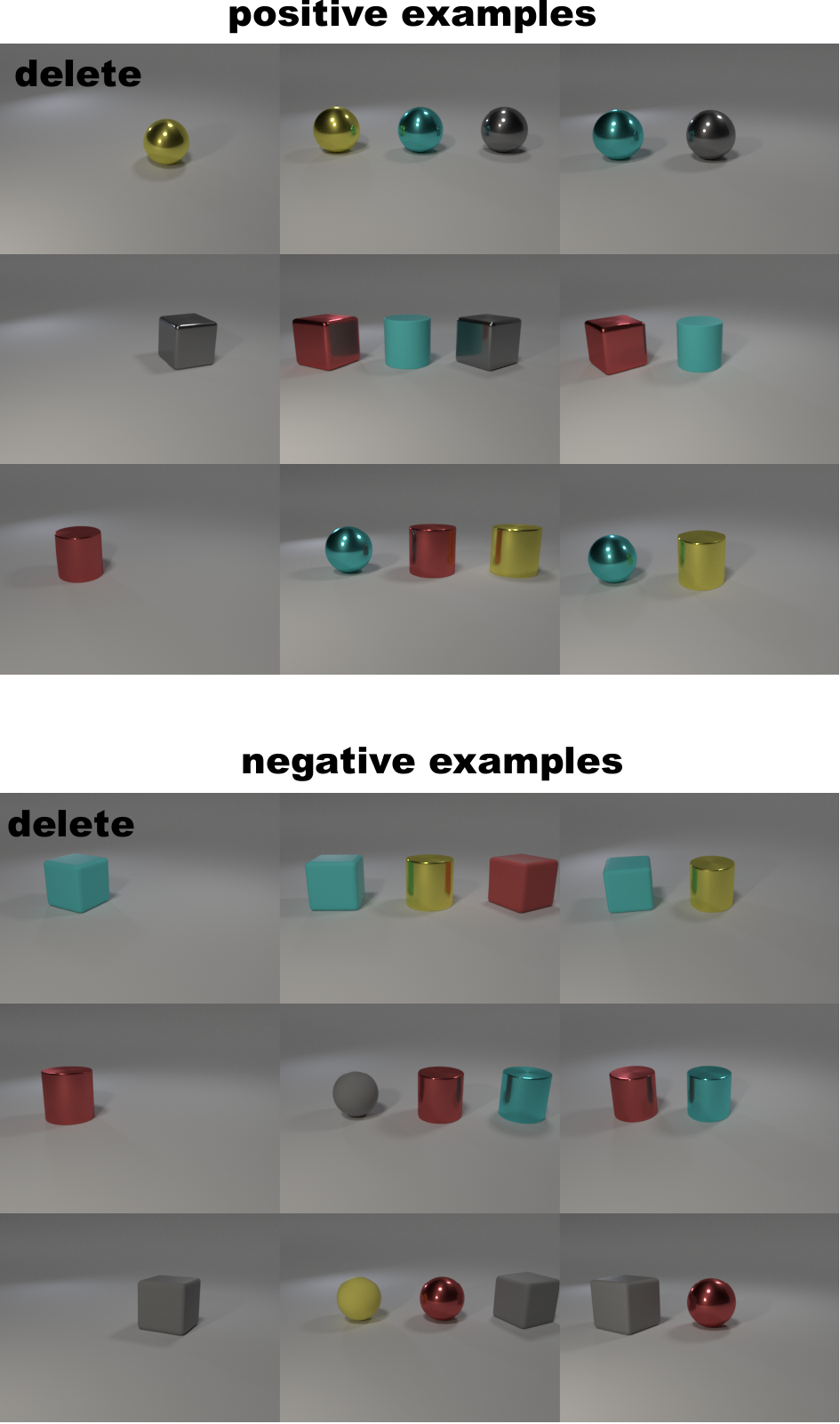}
    \caption{Positive and negative examples for \emph{delete} in CLEVR-List.}
    \label{fig:delete_examples}
\end{figure}

\begin{figure}[t]
    \centering
    \includegraphics[width=\linewidth]{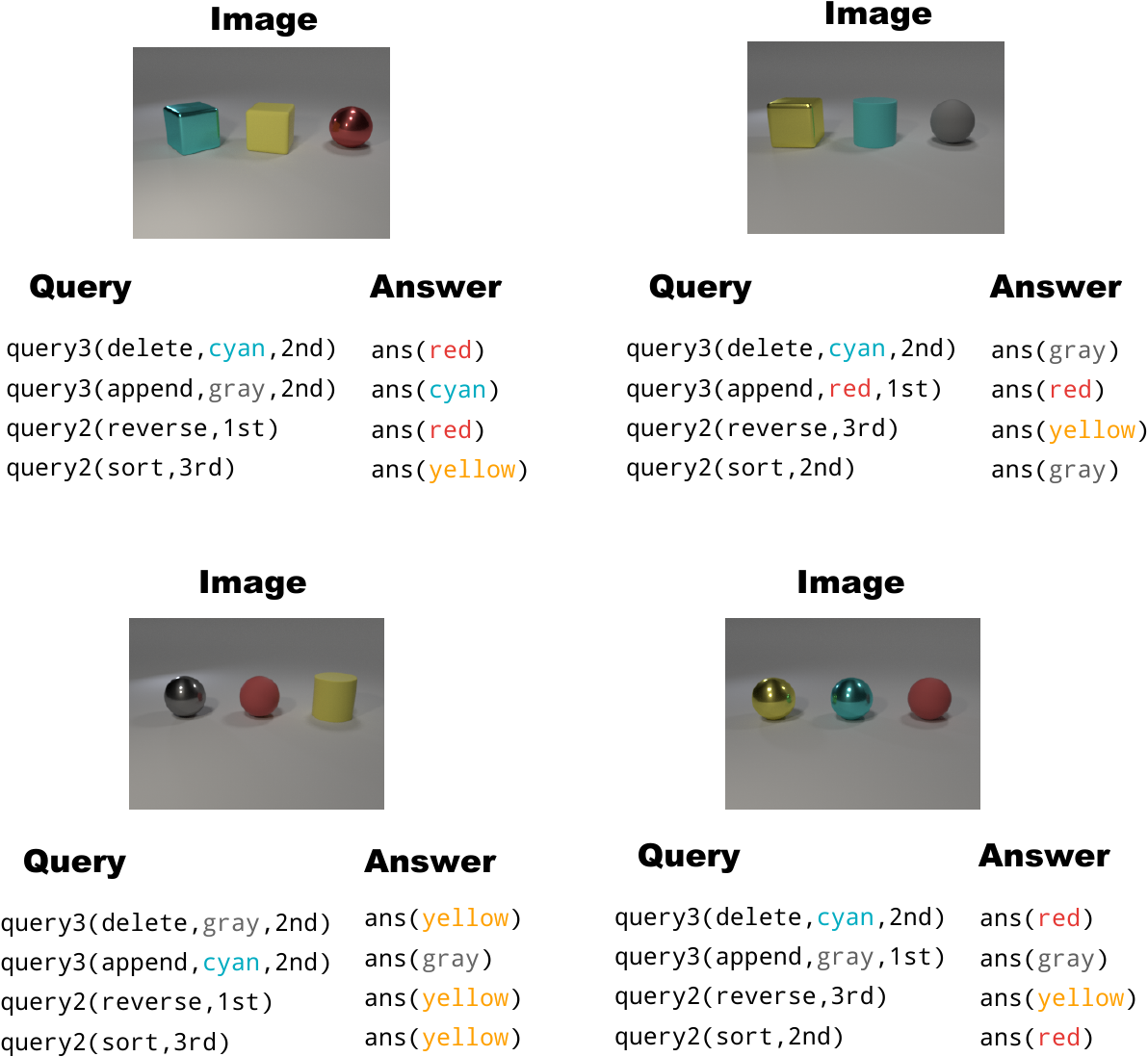}
    \caption{Examples of visual input, queries, and their answers in the Behind-the-Scenes task. Input is given as a pair of a visual scene and a query. }
    \label{fig:behind-the-scenes_examples}
\end{figure}

\begin{figure}[t]
\small
\begin{center}
      \begin{tabular}{c}
      \begin{minipage}{0.95\hsize}
      \centering
      \textbf{TwoPairs} \\
      {\em ``The
Kandinsky figure has two pairs of objects with the same shape. In one pair, the objects have the same colors in the other pair different colors. Two pairs are always disjunct, i.e., they do not share objects."}\\
      \begin{minipage}{0.24\hsize}
        \begin{center}
          \includegraphics[clip, width=\linewidth]{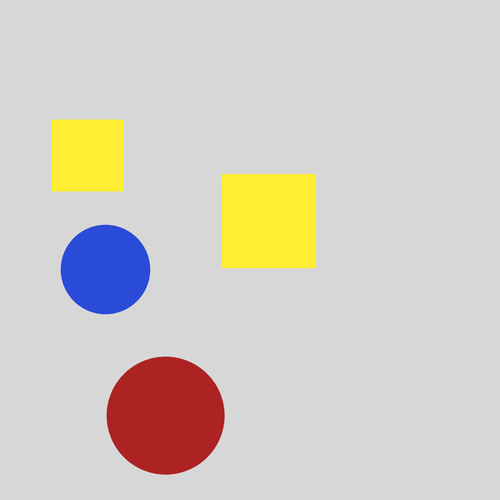}\\
          positive
        \end{center}
      \end{minipage}
      \begin{minipage}{0.24\hsize}
        \begin{center}
          \includegraphics[clip, width=\linewidth]{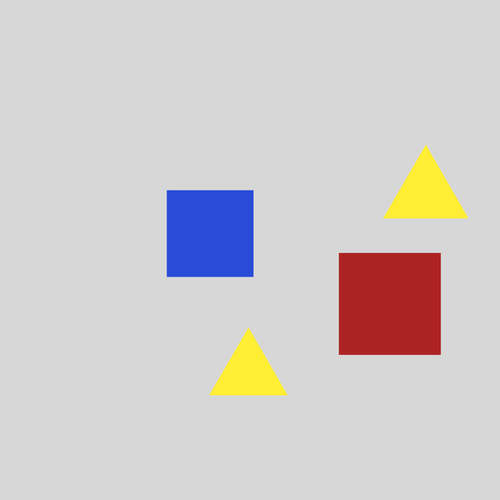}\\
          positive
        \end{center}
      \end{minipage}
            \begin{minipage}{0.24\hsize}
        \begin{center}
          \includegraphics[clip, width=\linewidth]{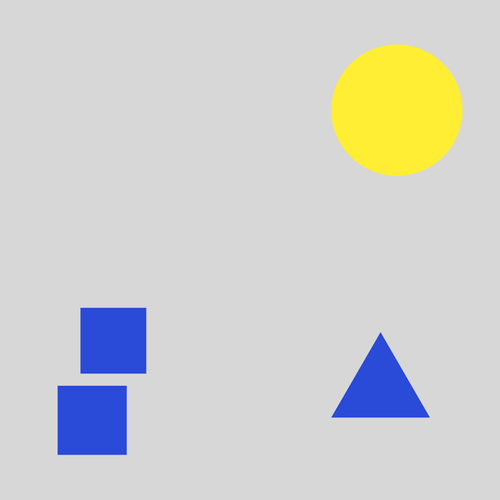}\\
          negative
        \end{center}
      \end{minipage}
      \begin{minipage}{0.24\hsize}
        \begin{center}
          \includegraphics[clip, width=\linewidth]{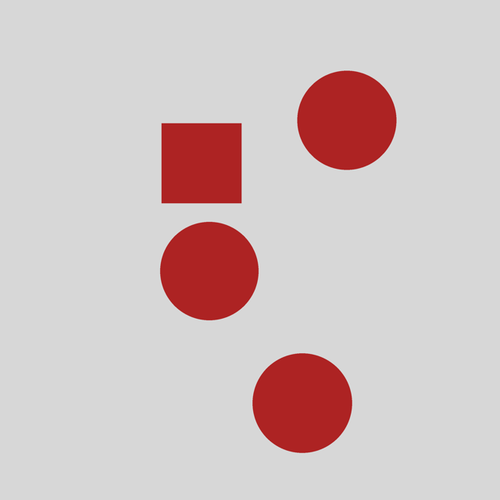}\\
          negative
        \end{center}
      \end{minipage}\vspace{.5em}\\
      \textbf{Closeby}\\
      {\em ``The
Kandinsky figure has a pair of objects that are close to each other."}.\\
        \begin{minipage}{0.24\hsize}
        \begin{center}
          \includegraphics[clip, width=\linewidth]{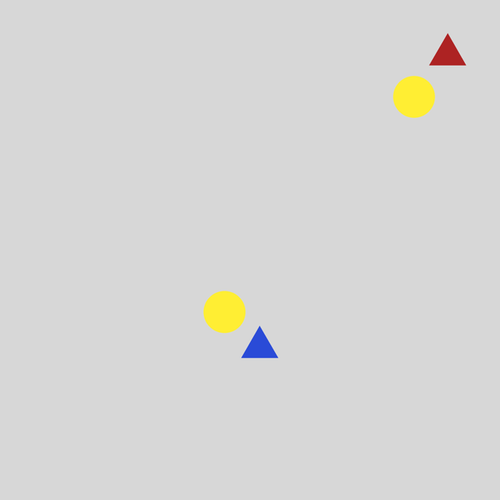}\\
          positive
        \end{center}
      \end{minipage}
              \begin{minipage}{0.24\hsize}
        \begin{center}
          \includegraphics[clip, width=\linewidth]{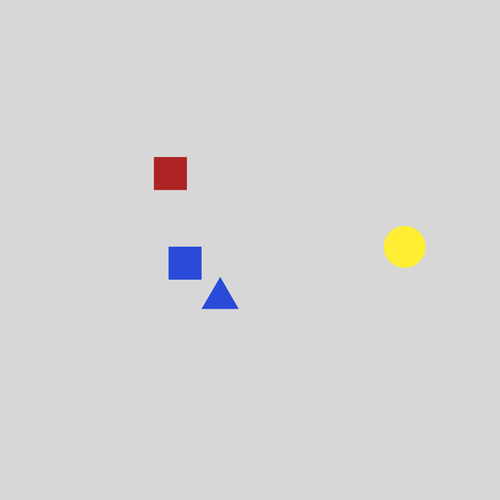}\\
          positive
        \end{center}
      \end{minipage}
      \begin{minipage}{0.24\hsize}
        \begin{center}
          \includegraphics[clip, width=\linewidth]{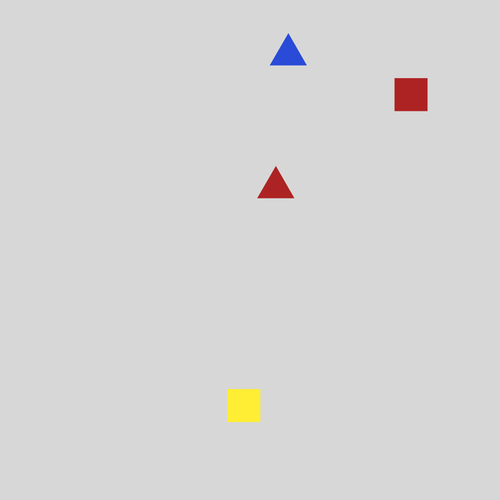}\\
          negative
        \end{center}
      \end{minipage}
              \begin{minipage}{0.24\hsize}
        \begin{center}
          \includegraphics[clip, width=\linewidth]{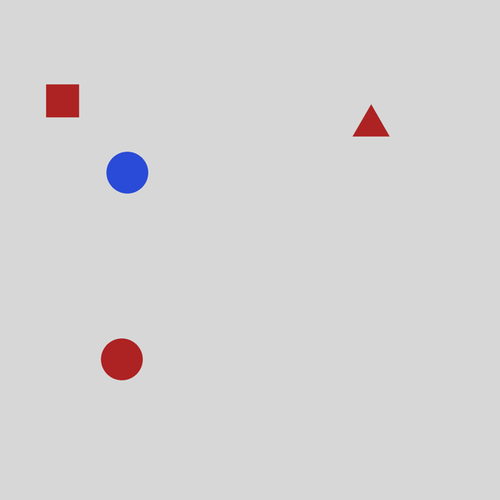}\\
          negative
        \end{center}
      \end{minipage}
    \vspace{.5em} \\
    \textbf{Red-Triangle}\\
    {\em ``The
Kandinsky figure has a pair of objects that are close to each other, and one object of the pair is a red triangle, and the other object has a different color and different shape. (A red triangle is attacking someone who has a different color and a different shape."}\\
              \begin{minipage}{0.24\hsize}
        \begin{center}
          \includegraphics[clip, width=\linewidth]{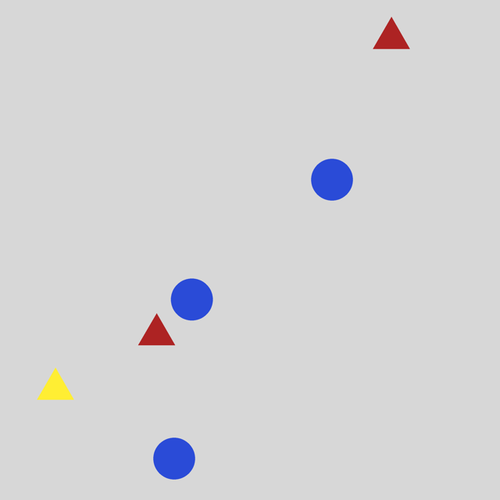}\\
          positive
        \end{center}
      \end{minipage}
              \begin{minipage}{0.24\hsize}
        \begin{center}
          \includegraphics[clip, width=\linewidth]{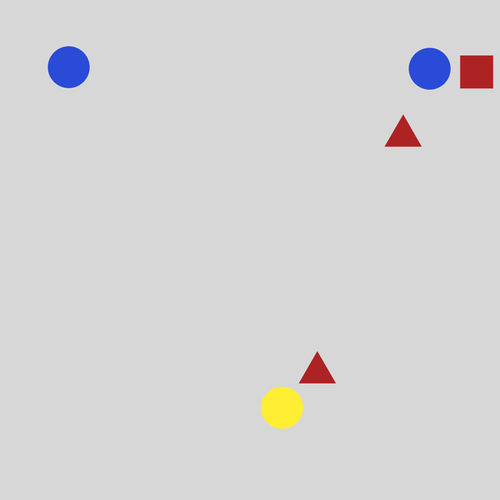}\\
         positive
        \end{center}
      \end{minipage}
      \begin{minipage}{0.24\hsize}
        \begin{center}
          \includegraphics[clip, width=\linewidth]{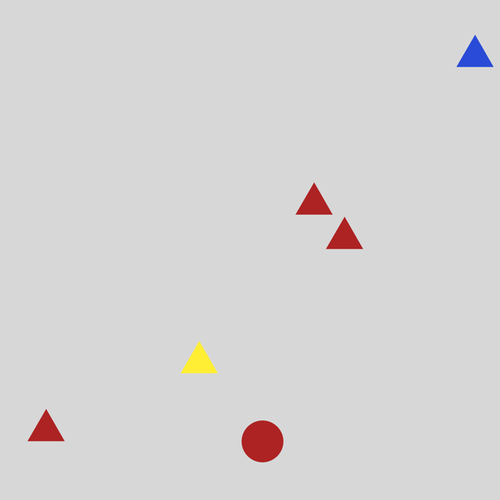}\\
          negative
        \end{center}
      \end{minipage}
              \begin{minipage}{0.24\hsize}
        \begin{center}
          \includegraphics[clip, width=\linewidth]{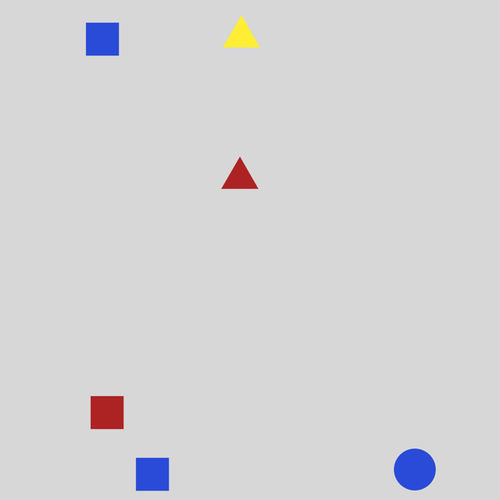}\\
          negative
        \end{center}
      \end{minipage}
     \vspace{.5em}\\
      \textbf{Online/Pair}\\
      {\em ``The
Kandinsky figure has five objects that are aligned on a line, and it contains at least one pair of objects that have the same shape and the same color."}\\
                 \begin{minipage}{0.24\hsize}
        \begin{center}
          \includegraphics[clip, width=\linewidth]{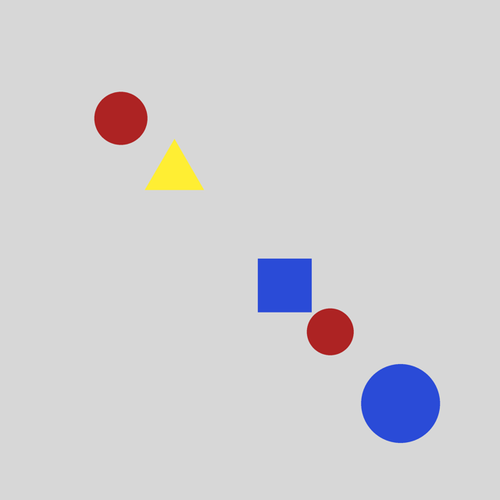}\\
          positive
        \end{center}
      \end{minipage}
      \begin{minipage}{0.24\hsize}
        \begin{center}
          \includegraphics[clip, width=\linewidth]{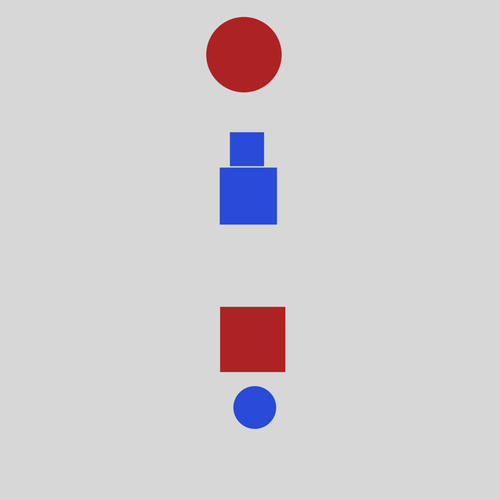}\\
          positive
        \end{center}
      \end{minipage}
        \begin{minipage}{0.24\hsize}
        \begin{center}
          \includegraphics[clip, width=\linewidth]{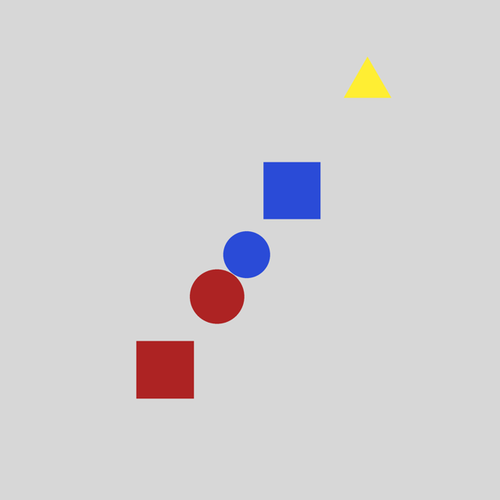}\\
          negative
        \end{center}
      \end{minipage}
            \begin{minipage}{0.24\hsize}
        \begin{center}
          \includegraphics[clip, width=\linewidth]{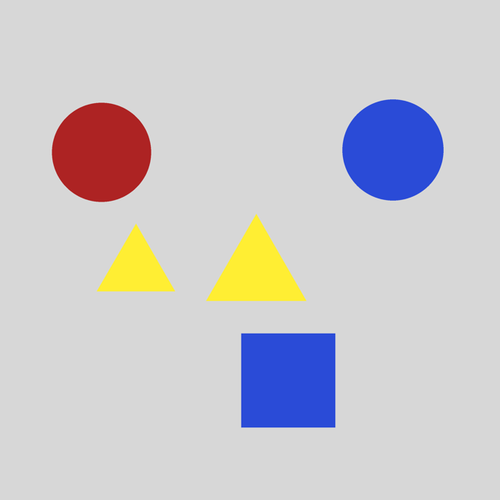}\\
          negative
        \end{center}
      \end{minipage}
 \end{minipage}
    \end{tabular}
    \end{center}
    \caption{Examples in each Kandinsky pattern in our experiments. The left two images are positive examples, and the right two images are negative examples.}
    \label{fig:kandinsky_examples}
\end{figure}

\begin{figure}[t]
\begin{center}
      \begin{tabular}{c}
      \underline{\textbf{CLEVR-Hans3}}\\
      \begin{minipage}{0.4\hsize}
        \begin{center}
        \textbf{class1}\\
            large (gray) cube\\
            and\\
            large cylinder
          \includegraphics[clip, width=\linewidth]{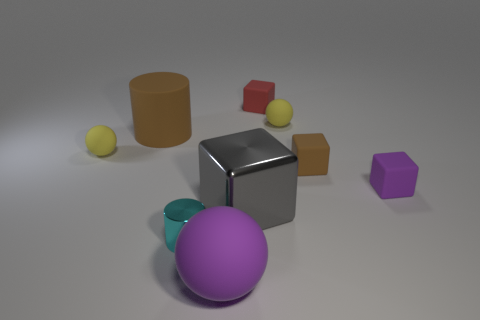}\\
          \includegraphics[clip, width=\linewidth]{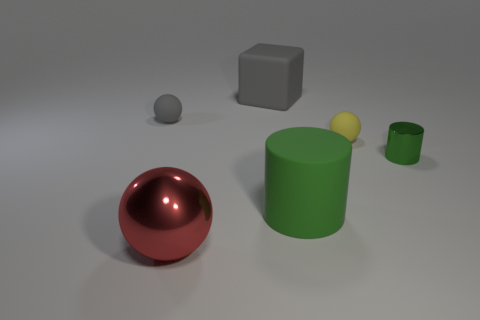}
        \end{center}
      \end{minipage}
      \begin{minipage}{0.4\hsize}
        \begin{center}
        \textbf{class2}\\
        small metal cube\\
        and\\
        small (metal) sphere
          \includegraphics[clip, width=\linewidth]{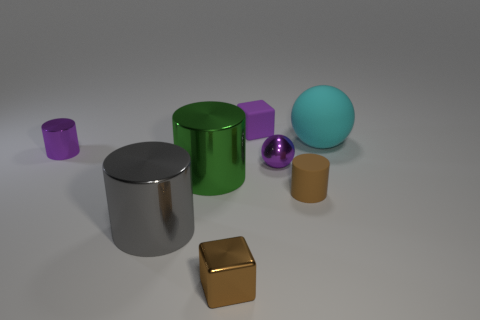}\\
          \includegraphics[clip, width=\linewidth]{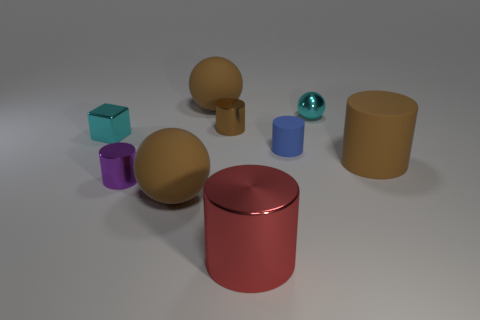}
        \end{center}
      \end{minipage}\vspace{.3em}\\
            \begin{minipage}{0.4\hsize}
        \begin{center}
        \textbf{class3}\\
        large blue sphere\\
        and\\
        small yellow sphere
          \includegraphics[clip, width=\linewidth]{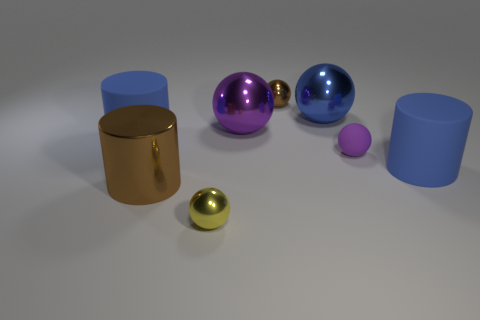}\\
          \includegraphics[clip, width=\linewidth]{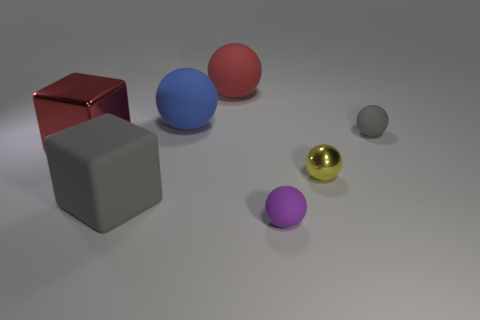}
        \end{center}
      \end{minipage}
    \end{tabular}
    \end{center}
    \caption{Examples in CLEVR-Hans3 dataset~\cite{Stammer21}. The dataset consists of \emph{three} classes. Two images are shown for each class. The text on the top of the images describes the confounded classification rule for each class. For example, images of the first class contain a large cube and a large cylinder. The large cube has the color of gray in every image of the train and validation split. In the test split, the color of the large cube is shuffled randomly.}
    \label{fig:clevr_hans_examples}
\end{figure}

Fig.~\ref{fig:member_sort_examples} shows some positive and negative examples for \emph{member} and \emph{sort} in CLEVR-List.
Fig.~\ref{fig:delete_examples} shows some positive and negative examples for \emph{delete} in CLEVR-List.
We show some examples of visual input, queries, and their answers in the behind-the-scenes task in Fig.~\ref{fig:behind-the-scenes_examples}.
We show some examples for each pattern we used in Kandinsky patterns in Fig. \ref{fig:kandinsky_examples}.
We also show some examples for each class of CLEVR-Hans in Fig. \ref{fig:clevr_hans_examples}.

\end{document}